\pdfoutput=1

\documentclass[11pt]{article}

\usepackage{pdfpages}

\usepackage[]{acl}

\usepackage{times}
\usepackage{latexsym}

\usepackage{multirow}
\usepackage{makecell}
\usepackage{adjustbox}
\usepackage{float}
\usepackage{xspace}
\usepackage{amsmath}

\usepackage[T1]{fontenc}

\usepackage[utf8]{inputenc}

\usepackage{microtype}

\usepackage{inconsolata}

\usepackage{graphicx}

\usepackage{booktabs}
 \usepackage{colortbl}

\newcommand{\INSQ}{{\sc debate}\xspace}
\newcommand{\RTRP}{{\sc panel}\xspace}

\title{WHoW: A Cross-domain Approach for Analysing Conversation Moderation}

\author{Ming-Bin Chen \and Lea Frermann \and Jey Han Lau \\
 School of Computing and Information Systems, The University of Melbourne\\
        \texttt{\{mingbin, lfrermann, laujh\}@unimelb.edu.au}}

\begin{document}
\maketitle
\begin{abstract}

We propose WHoW, an evaluation framework for analyzing the facilitation strategies of moderators across different domains/scenarios by examining their motives (\textbf{W}hy), dialogue acts (\textbf{Ho}w) and target speaker (\textbf{W}ho). Using this framework, we annotated 5,657 moderation sentences with human judges and 15,494 sentences with GPT-4o from two domains: TV debates and radio panel discussions. Comparative analysis demonstrates the framework's cross-domain generalisability and reveals distinct moderation strategies: debate moderators emphasise coordination and facilitate interaction through questions and instructions, while panel discussion moderators prioritize information provision and actively participate in discussions. Our analytical framework works for different moderation scenarios, enhances our understanding of moderation behaviour through automatic large-scale analysis, and facilitates the development of moderator agents.\footnote{Our code, dataset are available at \url{https://github.com/mrknight21/conversation_moderation_analysis}.}
\end{abstract}

\section{Introduction}
Conversational moderation typically involves a moderator who upholds an impartial stance and interest, to facilitate and coordinate discussions among participants through conversation~\citep{wright2009role}. Moderation occurs in diverse human interactive settings, however, the role of the moderator varies from hosts of debates~\citep{thale1989london, zhang2016conversational}, judges in judicial processes~\citep{danescu2012echoes}, to therapists in group therapy sessions~\citep{jacobs1998group}.

While there are various definitions of moderation across different domains~\citep{grimmelmann2015virtues, vecchi2021towards, friess2015systematic, trenel2009facilitation} the concept is generally characterized as a form of discourse optimization mechanism with the essential objectives of: (1) mitigation: preventing and policing negative behaviors, such as personal attacks~\citep{gorwa2020algorithmic}; (2) facilitation: promoting positive and constructive results, such as knowledge generation and consensus building~\citep{vasodavan2020moderation}; and (3) participation: ensuring balance and open participation opportunities for all members~\citep{kim2020bot}.

\begin{figure}[t]
\includegraphics[width=\columnwidth]{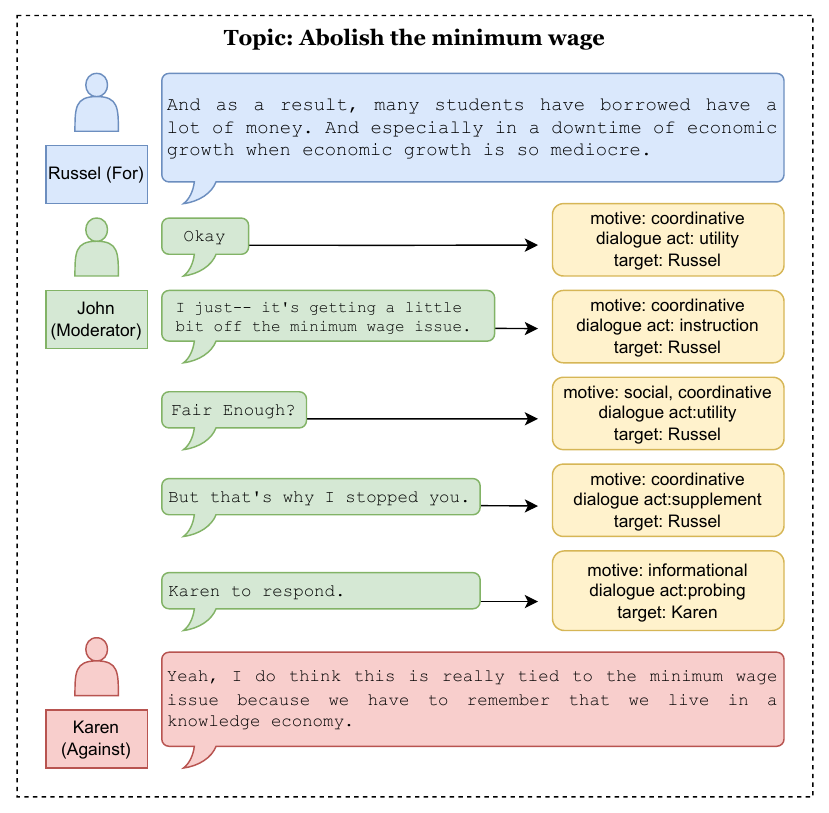}
\centering
\caption{Example of a moderated conversation and annotation using the WHoW framework. Blue, green, and red colors represent the supporting team, moderator, and opposing team in one of the \INSQ subset conversation, respectively. The peach-colored boxes contain the annotations for the corresponding moderator sentences.}
\label{fig:example_moderation}
\end{figure}

Extensive research has focused on content moderation analysis and automation in online spaces, primarily aimed at mitigating negative behaviors and intervening through asychronous actions such as post deletion~\citep{gorwa2020algorithmic, park2021detecting, wulczyn2017ex, falk2024moderation}. However, there are few studies that examine how moderators facilitate positive outcomes and balance participation through conversational engagement; our study seeks to address this.

We introduce WHoW: an analytical framework that breaks down the moderation decision-making process into three key components: motives (\textbf{W}hy), dialogue acts (\textbf{Ho}w), and target speaker (\textbf{W}ho). Using this framework, we analyzed transcripts of conversation moderation in two distinct domains: Intelligent Squared TV Debate (\INSQ) and Roundtable Radio Panel (\RTRP). We began by annotating 50 episodes with human annotators, which are then used to create and evaluate prompts for GPT-4o~\citep{openai2024api} for automatic annotation. We then used GPT-4o to automatically annotate more data and compared the moderation strategies in facilitating and balancing participation in these two domains.
Our findings reveal distinct moderation strategies and have the potential to support moderator training/assessment and facilitate the future development of moderator agents.

To summarise, our key contributions are:
\begin{enumerate}
\item We develop an analytical framework that characterizes conversational moderation across different scenarios using three dimensions: motives (Why), dialogue acts (How), and target speaker (Who);

\item Based on the framework, we annotated moderated multi-party conversations in two domains: TV debates and radio panel discussions. Our dataset comprises a total of 5,657 human-annotated sentences and model-annotated 15,494 sentences (GPT-4o).

\item By analyzing these two conversational domains—debates and panel discussions—we demonstrate the framework's cross-domain generalizability and identify distinct moderation strategies. Debate moderators focus on coordination, facilitating interactions through follow-up and confrontational questions, as well as instructions. In contrast, panel discussion moderators actively engage in and contribute to the topic themselves, while being less involved in fostering interactions between speakers.

\end{enumerate}

\section{Related Work} 
\label{subsec:relatedwork}

Conversation moderation is a complex task that requires consideration of multiple dimensions when making intervention decisions. This task takes place in multi-party settings~\citep{gu2021mpc, ganesh2023survey}, where a moderator's decisions regarding interventions and turn assignment~\citep{hyden2003s, gibson2003participation, ouchi2016addressee, wei2023multi} must account for the conversation context, group dynamics, and the balance of participation. Depending on the scenario, moderators fulfill various functional roles, such as inviting for contribution, providing background information, facilitating topic transitions, and posing questions to guide discussions and maintain their quality~\citep{wright2009role, park2012facilitative, mao2024multi, schroeder2024fora}. Furthermore, moderators often operate under hybrid motives, which include facilitating quality arguments~\citep{landwehr2014facilitating}, maintaining social engagement~\citep{myers2014becoming}, and managing external factors like time constraints~\citep{wright2009role}. Ultimately, moderation is a strategic task, requiring the application of specific strategies to encourage constructive contributions and participant engagement while minimizing destructive conflicts~\citep{hsieh2012effect, edwards2002moderator, forester2006making}.

The effect and influence of moderation have been studied across various domains using different analytical measures. In online mental health support forums, the presence of a moderator has been shown to improve user engagement, openness, linguistic coordination, and trust-building compared to non-moderated groups~\citep{wadden2021effect}. In the educational domain, moderators have been found to enhance collaboration patterns and increase online participation rates in group learning settings~\cite{hsieh2012effect}. Case studies and interviews have also been conducted to analyze the role and function of moderators in community building~\cite{cullen2022practicing, seering2019moderator}, focus group discussions~\cite{gronkjaer2011analysing}, online public issue discussions and debates~\cite{wright2009role, edwards2002moderator}, and mediating contentious stakeholders~\cite{forester2006making}.

Despite the existence of some annotation protocols and datasets, resources for conversational moderation remain notably limited. Many studies have been conducted on small sample sizes~\cite{vasodavan2020moderation, hsieh2012effect} and often do not make their datasets publicly available~\cite{gronkjaer2011analysing, wadden2021effect}. Additionally, the research often relies on methodologies such as interviews or case studies, which are not reusable for further analysis or automation~\cite{forester2006making}. As of the time this study's manual annotation was conducted, the only annotated dataset currently available consists of just 300 comments~\citep{park2012facilitative}\footnote{In concurrent work, \citet{schroeder2024fora} published a substantial facilitative conversation corpus.}. Furthermore, some studies treat moderation as a reactive intervention to participant comments and structure the data as comment-intervention pairs~\citep{falk2024moderation, gronkjaer2011analysing}, thereby overlooking broader session-level objectives such as balancing participation and the overall role of the moderator. Moreover, while several annotation protocols exist, they tend to be overly specific to their application domains. For instance, the role of ``resolving site use issues'' is only pertinent to e-rule-making scenarios~\cite{park2012facilitative} and don't generalise to other domains.

\section{The WHoW Conversational Moderation Analytic Framework}
\label{sec:framework}

\begin{table*}[]
\resizebox{\textwidth}{!}{%
\begin{tabular}{p{0.115\linewidth} p{0.185\linewidth} p{0.785\linewidth}}
\toprule
{\bf Dimension} & {\bf Label} & {\bf Definition} \\ \midrule
\multirow{6}{*}{Motives} & Informational (IM) & Provide or acquire relevant information to constructively advance the topic or goal of the conversation. \\
 & Coordinative (CM) & Ensure adherence to rules, plans, and broader contextual constraints, such as time and environment. \\
 & Social (SM) & Enhance the social atmosphere and connections among participants by addressing feelings, emotions, and interpersonal dynamics within the group. \\ \midrule
\multirow{11}{2cm}{Dialogue acts} & Probing (prob) & Prompt speaker for responses.\\  \\
 & Confronting (conf) & Prompt one speaker to response or engage with another speaker's statement, question or opinion. \\
 & Instruction (inst) & Explicitly command, influence, halt, or shape the immediate behavior of the recipients. \\
 & Interpretation (inte) & Clarify, reframe, summarize, paraphrase, or make connection to earlier conversation content. \\
 & Supplement (supp) & Enrich the conversation by supplementing  details or information without immediately changing the target speaker's behavior. \\
 & Utility (util) & All other unspecified acts. \\ \midrule
Target speaker & Target speaker (TS) & The group or person addressed by the moderator. \\ \bottomrule
\end{tabular}%
}
\caption{Definitions and acronyms for the labels across the three dimensions: motives (Why), dialogue acts (How), and target speakers (Who). Target Speaker is a categorical variable with values corresponding to each participant in the dialogue, plus ``audience'', ``self'', ``everyone'', ``support side'', ``against side'', ``all speakers'', and ``unknkown''.}
\label{tab:def}
\end{table*}

We design an analytical framework that: (1) is grounded in the existing literature~\citep{wright2009role, park2012facilitative, vasodavan2020moderation, lim2011critical}; (2) captures the multifaceted nature of conversational moderation; and (3) generalizes across different domains. Our framework (Table~\ref{tab:def}), inspired by existing multi-party agent work~\citep{wei2023multi, mao2024multi}, is structured around three core dimensions: motives (Why), dialogue acts (How) and target speakers (Who). In addition to dialogue acts, which are widely employed to study dialogue patterns~\citep{shriberg2004icsi}, we incorporate the motive dimension to provide insights into the \textit{reason} the moderator intervenes given a particular scenario or context~\citep{yeomans2022conversational}. Furthermore, we introduce the target speaker dimension to explore the moderator’s interactive style and strategies for balancing participation in a multi-party setting~\citep{gibson2003participation, hyden2003s}. By decomposing the moderation process into these distinct components and analyzing their interplay, the framework enables the characterization of moderator behavior.

Table \ref{tab:def} shows the label definitions under the three dimensions. To derive our labels and understand their compatibility with existing protocols, we categorize all moderation-related typologies identified in Section~\ref{subsec:relatedwork} into motives and dialogue acts, as detailed in Appendix Table \ref{tab:gen}. Since moderator responses can be lengthy, and may serve multiple goals (i.e., correspond to multiple labels), we first break them into individual sentences and then label each sentence across the three dimensions using the definitions provided above. We elaborate on these three dimensions in the following sections.

\subsection{Motives: Why does the moderator intervene?}

The ``Why'' component examines the motivations behind a moderator's interventions in conversations. Existing protocols distinguish socially motivated speech --- such as ``affective strategy''~\citep{hsieh2012effect} and ``social functions''~\citep{park2012facilitative} --- from argument-driven speech. This pattern aligns with the conversational circumplex framework, which categorizes conversational goals along informational and relational dimensions~\citep{yeomans2022conversational}. Furthermore, in facilitated group debates like Intelligent Squared Debate~\citep{zhang2016conversational} moderator interventions can be motivated by meeting rules, such as adherence to time limits. Consequently, we propose three motives driving moderation behaviors: informational, social, and coordinative (Table \ref{tab:def}, top), which align with the facilitation types described by \citet{lim2011critical} and accommodate the hybrid-motive nature of the moderator role. As previous studies~\citep{yeomans2022conversational} and our pilot studies show that a single speech can convey multiple motives, we treat this annotation as a multi-label task (e.g., a moderator sentence may have both social and coordinative motives). 

\subsection{Dialogue Acts: How does the moderator intervene?}

The ``How'' component analyzes the dialogue acts, or the immediate functions of a moderator's interventions. By examining the sequential patterns of these acts, we gain insights into the strategies moderators use to realize their motives. The initial set of dialogue acts is derived from the five fundamental labels of the MRDA corpus~\citep{shriberg2004icsi}, which was developed for annotating multi-party meetings: ``Question', ``Statement'', ``BackChannel'', ``Disruption'', and ``FloorGrabber''. The two major labels, ``Question'', and ``Statement'', indicate the functions of information elicitation and information provision respectively. These two major labels are instrumental in distinguishing the moderator's functional role as either a ``Interviewer'' or an ``Contributor'' respectively~\citep{mclafferty2004focus}. The remaining MRDA labels, along with other unspecified acts, such as greeting, are grouped into the 'Utility' category, as they do not directly contribute to information exchange."

We further categorise the two major labels into sub-labels to capture the nuanced characteristics of moderators' interventions. For ``Question'', we distinguish two types of information elicitation interventions to capture whether the moderator seeks to acquire information through direct prompts (\textbf{Probing}) or by encouraging interaction among participants (\textbf{Confronting}). Turning to ``Statement'', we distinguish between interjections that change a participant's behavior (e.g.\ command to stop; \textbf{Instruction}), refer back to prior discussion (e.g.\ summarization; \textbf{Interpretation}), and provide additional information or opinions (\textbf{Supplement}) \citep{park2012facilitative, wright2009role}.

The detailed definitions of these fine-grained labels are included in Table \ref{tab:def} (middle). Appendix Table \ref{tab:example_sents} presents example sentences that intersect between the motives and dialogue acts dimensions. We treat dialogue acts as mutually exclusive and formalize it as a multi-class classification task.

\subsection{Target Speaker: Who does the moderator address?}

The ``Who'' component focuses on identifying the intended target of the moderator's intervention, which differs from the typical task of next-speaker prediction in multi-party dialogues~\citep{ishii2019prediction}. Since the target participants are not always the subsequent speakers, analyzing the discrepancies between the prior speaker, target speaker, and next speaker allows for an assessment of the intended shifts in participation and the moderator's initiatives during the discussion.

We approach the annotation of this dimension as a multi-class classification task, with labels corresponding to speakers. To accommodate different contexts, we also introduce general labels such as ``everyone'' (including audience), ``unknown'', and ``all speakers''. For the TV debates domain specifically we introduce 3 additional labels ``audience'', ``against team'', and ``support team''. While our framework is designed to be cross-domain, we note that its labels or categories are customizable depending on the domain.

\section{Dataset and Human Annotation}

\begin{table}[t]
\centering
\resizebox{\linewidth}{!}{
\begin{tabular}{ll@{\;}l@{\;}ll@{\;}l}
\toprule
 & \multicolumn{3}{c}{\bf \INSQ}                            & \multicolumn{2}{c}{\bf \RTRP}          \\ \cline{2-6} 
                         & {\bf Test}            & {\bf Dev }            & {\bf Train}           & {\bf Test}            & {\bf Train}           \\ \hline
Episodes           & 19              & 11              & 78              & 20              & 68 \\
Speakers / episode Mean          & 4.63   & 4.55  & 4.62  & 3.450  & 4.47   \\\midrule
M Share / episode (\%) & 38\% & 36\% & 37\% & 41\% & 40\% \\
M Turns / episode    & 69  & 73  & 70  & 17  & 21  \\
M Sentences (Total)  & 2,795            & 1,702            & 11,153           & 1,160            & 4,341            \\ \bottomrule
\end{tabular}
}
\caption{Descriptive statistics for the \INSQ and \RTRP. M denotes Moderator; share the proportion of words uttered by the moderator; and turn the full utterance (which contains multiple sentences).}
\label{tab:dataset_stats}
\end{table}

\subsection{Datasets}

We use the Intelligence Squared Debates  Corpus~\citep{zhang2016conversational} (henceforth \INSQ), a collection of transcripts from a live-recorded U.S. television debate show featuring Oxford-style debates. The corpus comprises 108 episodes covering a wide range of topics, from foreign policy to the benefits of organic foods. Each debate includes a moderator and two teams of experts arguing, respectively, ``for'' and ``against'' the topic. Although the debates are structured into three phases—introduction, discussion, and conclusion—our analysis focused exclusively on the interactive discussion phase, where the majority of the moderated interactions occur.\footnote{In addition, the corpus provides information on each speaker's role (moderator, team member, audience member) and metadata, such as short bios and audience voting results before and after the debate.} We randomly split the episodes into 11 for development, 19 for testing, and 78 for training.

To validate the generalizability of our framework across scenarios, we also include a second dataset from a subset of The NPR Interview Corpus~\citep{majumder2020interview} (henceforth \RTRP). We specifically select episodes from a panel discussion program titled ``Roundtable'', in which the moderator accounts for 30\% - 50\% of the dialogue, and which involve more than three speakers. This subset features panel discussions with speakers holding diverse views, though not necessarily opposing each other (unlike \INSQ). This selection yielded 88 episodes, from which we randomly sampled 20 episodes to create a test set. Table \ref{tab:dataset_stats} presents some descriptive statistics of the two datasets.

\subsection{Human annotation process}

We recruited annotators to label each sentence of the moderator's utterance based on the WHoW framework, as illustrated in Figure \ref{fig:example_moderation}.\footnote{The development of the annotation schema started with two rounds of pilot studies involving the paper authors and NLP PhD students for testing the preliminary  definitions of the framework with one episode from each dataset. Feedback from the pilot resulted in framing the motive labels as a multi-label task and reducing the dialogue act classes from eight to six with refined definitions.}
\begin{table}[t]
\centering
\begin{tabular}{lccccc}
\toprule
 & \textbf{DA} & \textbf{IM} & \textbf{CM} & \textbf{SM} & \textbf{TS} \\
\midrule
\INSQ & 0.49 & 0.43 & 0.37 & 0.41 & 0.72 \\
\RTRP     & 0.59 & 0.67 & 0.54 & 0.63 & 0.75 \\
\bottomrule
\end{tabular}
\caption{Inter-annotator agreement (Krippendorff's alpha), across the dialogue acts (DA), motives (IM, CM, SM), and target speaker (TS) dimensions for the datasets \INSQ and \RTRP.}
\label{tab:agreement}
\end{table}
We recruited five annotators in total, all proficient or native English speakers and students of either linguistics or NLP, and they were paid 36 USD/hour. The annotators manually annotated the development and test sets of \INSQ and the test set of \RTRP. Annotators received the definitions of labels as outlined in Section \ref{sec:framework} and Table \ref{tab:def}. To facilitate the dialogue act annotation and increase agreement, we developed a decision tree flowchart (see Appendix Figure \ref{fig:da_tree}). We conducted one practice annotation round including group discussions to clarify any misconceptions and two further meetings during the annotation phase to discuss remaining misunderstandings. More details of annotation material and interface are provided in Appendix Section~\ref{sec:anno}.

Each sentence in the moderators' utterance was annotated for the presence of the three motives, one identified dialogue act, and the target speaker(s). Each episode was annotated by at least two annotators. The final ground truth were aggregated using majority vote; in cases of evenly divided annotation votes, the first author did the tie-breaking. Inter-annotator agreement (Krippendorff's alpha) is presented in Table \ref{tab:agreement}. \RTRP generally has higher agreement and the overall agreement ranges from moderate to good and these numbers are consistent with previous studies that involve complex and subjective judgements~\citep{falk2024moderation}. A detailed analysis of disagreements is provided in Appendix section \ref{sec:disagree}.

\section{Automatic Annotation}
\label{sec:autoanno}
\begin{table}
\setlength\tabcolsep{1.5pt}
\centering
\begin{adjustbox}{max width=\linewidth }
\begin{tabular}{lccccc}
\toprule
\textbf{Model} & \textbf{DA} & \textbf{IM} & \textbf{CM} & \textbf{SM} & \textbf{TS} \\
\midrule
Random(\INSQ)        & 0.153          & 0.492                   & 0.508            & 0.405                  & 0.057             \\
MT(\INSQ)                 & 0.485          & \textbf{0.761}                  & \textbf{0.711}          & \textbf{0.767}                 & 0.497             \\
ST(\INSQ)     & \textbf{0.515}          & 0.7287                  & 0.686           & 0.668                 & \textbf{0.525}            \\
\midrule
Random(\RTRP)        & 0.115          & 0.490                   & 0.482            & 0.387                  & 0.096 \\
MT(\RTRP)                 & \textbf{0.504}          & 0.726                  & \textbf{0.732}           & \textbf{0.754}                 & \textbf{0.467}             \\
ST(\RTRP)     & 0.492          & \textbf{0.747}                  & 0.639           & 0.635                 & 0.464            \\

\bottomrule
\end{tabular}
\end{adjustbox}
\caption{Macro-F1 comparing GPT-4o using multi-task (MT) and single-task (ST) approaches across the two subsets. The bold numbers highlights the top performer of the dimension in the subset. The random baseline are derived from five random simulations. }
\label{tab:class_result}
\end{table}

\begin{table}
\centering
\begin{adjustbox}{max width=\linewidth }
\begin{tabular}{lccccc}
\toprule
\textbf{Model} & \textbf{DA} & \textbf{IM} & \textbf{CM} & \textbf{SM} & \textbf{TS} \\ \midrule
MT (\INSQ) & 0.38 & \textbf{0.52} & \textbf{0.42} & \textbf{0.53} & 0.66 \\
ST (\INSQ) & \textbf{0.53} & 0.46 & 0.37 & 0.34 & \textbf{0.68} \\ \midrule
MT (\RTRP) & \textbf{0.53} & 0.45 & \textbf{0.51} & \textbf{0.46} & 0.60 \\
ST (\RTRP) & {\bf 0.53} & \textbf{0.49} & 0.28 & 0.27 & \textbf{0.61} \\ \bottomrule
\end{tabular}
\end{adjustbox}
\caption{Krippendorff’s alpha agreement between the (majority) human labels and GPT-4o predictions using single task (ST) or multi-task (MT) prompts for the two datasets.}
\label{tab:llm_agreement}
\end{table}

Manual labeling is time-consuming and extensive. A practical and generalizable framework for large-scale exploration requires an automatic labeling framework. To this end, we leverage GPT-4o~\citep{openai2024api} for automatic annotation. We optimized the prompts using the development set from  \INSQ (see Appendix section \ref{sec:prompt}. for more details on the prompt design). Our single-task setting (``ST'') frames the annotation as five independent classification tasks: two multi-class classifications for dialogue acts (``DA'') and target speakers (``TS''), and three binary classifications for motive labels (``IM'', ``CM'', ``SM''). In addition to ST, we also developed a version of GPT-4o to perform all tasks jointly with one single prompt (multi-task or ``MT''). We present macro-F1 and agreement results of ST and MT with human test set annotations in Table~\ref{tab:class_result} and Table \ref{tab:llm_agreement} respectively. Overall, the results are encouraging and demonstrate that GPT-4o is a viable method for automatic annotation, particularly given the tasks' high level of subjectivity and complexity  (\citet{falk2024moderation}; Appendix Section \ref{sec:disagree}). Error analysis (Appendix Section \ref{sec:error}) reveals that most mis-classifications arise from subjective interpretations, context dependency, or ambiguity in extremely long or short sentences. As the multi-tasking approach (MT) has higher average Macro-F1 (0.64 vs.\ 0.61) and agreement (0.51 vs.\ 0.46) across tasks and datasets, we used this approach for automatic annotation, and ran it on the training sets of \INSQ and \RTRP.\footnote{We also experimented with using the automatically annotated training sets to fine-tune smaller supervised language models, such as Longformer~\citep{beltagy2020longformer},
and found it works quite well; see Appendix Section \ref{sec:longformer}.}

\section{Analysis}

We now analyze which dialogue acts are used to facilitate discussion and encourage participation. To demonstrate the generalizability of the automatic framework, this analysis draws on our full dataset, including development, test, and train sets, annotated using GPT-4o with the multi-task approach. In Appendix Section~\ref{sec:human_machine}, we compare GPT-4o labels against human label distributions on the development and test set, showing that they are overall consistent, with the exception of the "instruction" and "supplement" acts, with only minor variations in magnitude. Specifically, we examine how speaker rotation is facilitated and how the three motives are addressed across the two domains in the dataset.

\begin{table}[]
\setlength\tabcolsep{4.5pt}
\centering
\begin{small}
\begin{tabular}{@{}llllllll@{}}
\toprule
\multicolumn{8}{c}{\INSQ}\\ \midrule
      & prob                           & conf                         & inst                           & inte   & supp                                                  & util                           & $p(m)$                         \\ \midrule
IM    & \textbf{0.41} & \underline{0.23*} & 0.04*                         & 0.11* &  0.20 & 0.01                          & 0.39  \\
CM    & \underline{0.15*} & 0.10*                       & \textbf{0.54*} & 0.02 & 0.09                                                & 0.10                         & 0.66* \\
SM    & 0.08                          & 0.02                       & 0.10*                         & 0.02 & \underline{0.14}                      & \textbf{0.65} & 0.12                          \\\midrule
$p(d)$ & 0.22   & 0.11*                         & 0.36*   & 0.05*    & 0.12                                                  & 0.14*                           &  \\ \midrule
\multicolumn{8}{c}{\RTRP}                       \\ \midrule
IM    & \underline{0.42} & 0.03 & 0.01                         & 0.03 & \textbf{0.51*} & 0.01                         & 0.72* \\
CM    & 0.06 & 0.02                         & \textbf{0.42} & 0.01 & \underline{0.33*}                       & 0.17*                          & 0.25 \\
SM    & 0.05                         & 0.01                         & 0.02                         & 0.01 & \underline{0.28}                        & \textbf{0.63} & 0.16*                          \\\midrule
$p(d)$ & 0.30*   & 0.03                            & 0.10    & 0.02    & 0.41*                          & 0.13                            & \\ \bottomrule
\end{tabular}%
\end{small}
\caption{Conditional probabilities of dialogue acts ($d$; columns) given motives ($m$; rows), with marginal probabilities averaged across episodes for the two scenarios—\INSQ (top) and \RTRP (bottom). The most likely dialogue act per motive is highlighted in bold, and the second most likely is underlined. 
$^*$ indicates a significantly larger $p(d|m)$ in one data set compared to the other (t-test; p$<=0.05$). On average, a moderator speaks 151 sentences per \INSQ episode and 61 per \RTRP episode. }
\label{tab:insq_co}
\end{table}

\subsection{Motives and Dialogue Acts}

Table \ref{tab:insq_co} presents the probabilities that motive $m$ (rows) is realized by dialogue act $d$ (columns), $p(d|m)$, as well as all dialogue acts and motives labels' marginal probabilities.
There is a distinct difference in relative motive frequencies between the two domains. \INSQ moderation is dominated by a coordinative motive (66\%) followed by informational (39\%).
In contrast, informational motives are the most frequent in \RTRP moderation (72\%). 
For relative dialogue act frequencies, \INSQ moderators mostly focus on providing instructions (36\%), while \RTRP moderators tend to supply information (41\%). Probing is the second most common dialogue act in both corpora.

Turning to the conditional probabilities, strategically, \INSQ moderators achieve {\bf informational motives} (IM) by actively facilitating participant contributions through methods such as probing (0.41) and confronting (0.23), along with notable uses of interpretation (0.12) and supplementing (0.19) information. IM in \RTRP, on the other hand, is characterized by moderators delivering information themselves (0.51) and engaging participants through probing (0.41). The minimal use of confrontation (0.03) and interpretation (0.01) in \RTRP indicates relatively few attempts to foster interaction and engagement between non-moderator participants. 

Conversely, \INSQ moderators more frequently prompt participants to respond to one another (confronting) and engage with earlier discussion content (interpretation). Overall, for IM \INSQ moderators’ interventions are more diverse and leading to interaction between participants compared to those in \RTRP.

{\bf Coordination motives} (CM) in both domains primarily rely on instructions (0.54 in \INSQ and 0.42 in \RTRP). However, \INSQ moderators are more likely to coordinate through probing (0.15), maintaining dialogue engagement by asking participants about their preferences for rotation and participation. \RTRP moderators coordinate by providing supplementary information (0.33), e.g.\ by explaining rules. 

While \RTRP has a significantly higher proportion of moderator interventions driven by {\bf Social Motives} (SM) compared to \INSQ, there is no notable difference in the dialogue acts used. Both settings primarily utilize utility acts (0.65 in \INSQ, 0.62 in \RTRP), such as greetings, along with some social/personal information sharing (supplement) to fulfill their social motives. Although our observations can be partially explained by the respective rules of the discussion programs, they highlight different high-level strategies to facilitate a constructive discussion.

\subsection{Balancing Speaker Participation}

An essential role of a moderator is to facilitate balanced participation among participants and their respective stances. To analyze how moderators balance participation, we examine the transition probabilities between moderator dialogue acts and speaker rotation. 

Given an episode of a conversation consisting of \( n \) turns between a moderator and participants, we denote the speaker identities (e.g.\ moderator or name of a participant) as \( [p_{0}, p_{1}, \ldots, p_{n}] \). Note that a turn here denotes the full utterance (which can have multiple sentences) by a speaker.

\begin{table}[]
\centering
\begin{small}
\begin{tabular}{@{}lccc@{}}
\toprule
\multicolumn{4}{c}{\INSQ} \\ \midrule
 & moderation & continuation & rotation \\ \midrule
moderation & -- & 0.52 & 0.48 \\
continuation & 0.78 & -- & 0.22 \\
rotation & 0.47 & -- & 0.53 \\ \midrule
\multicolumn{4}{c}{\RTRP} \\ \midrule
moderation & -- & 0.35 & 0.65 \\
continuation & 0.80 & -- & 0.20 \\
rotation & 0.60 & -- & 0.40 \\ \bottomrule
\end{tabular}%
\end{small}
\caption{Transition probabilities between moderator interventions and speaker rotation / continuation. Note: the transition from 'rotation' to 'rotation' represents instances of participant-driven rotation without moderator intervention. `--' indicates that transitions are not possible.}
\label{tab:rotation_trans}
\end{table}
To understand the rotation pattern (i.e.\ how the dialogue transition from one speaker to another), we simplify the speaker status for each turn ($s_t$) as follows:

\[
\ s_t =
\begin{cases} 
\text{\it moderation}, &\text{if } p_t = \text{moderator} \\
\text{\it continuation}, &\text{if } p_t \neq \text{moderator} \And \\
&\phantom{xxx} p_t = p_{t'}\\
\text{\it rotation}, &\text{if } p_t \neq \text{moderator} \And \\
&\phantom{xxx} p_t \neq p_{t'}
\end{cases}
\]
where $t'$ is the last non-moderator turn before $t$.

By converting the conversation sequence into three states—moderation, continuation, and rotation—we derive a transition probability matrix ($P(s_{t+1}|s_{t})$), as shown in Table \ref{tab:rotation_trans}. The table reveals several key patterns: both \INSQ and \RTRP moderators are more likely to intervene when a speaker has continued for more than one exchange (0.78 and 0.80). However, \INSQ moderators (0.52) exhibit a higher tendency than \RTRP moderators (0.35) to continue the conversation with the same participant; or another interpretation is that \RTRP moderator intervention has a higher tendency to lead to speaker \textit{rotation}.
Additionally, there are more participant-driven rotations ($rotation \rightarrow rotation$) in the \INSQ dataset (0.53) compared to the \RTRP dataset (0.40), indicating a higher level of \textit{independent} interaction among participants in \INSQ.

\begin{figure}[t!]
\includegraphics[width=\columnwidth,clip,trim=1cm 0.5cm 0.5cm 1cm]{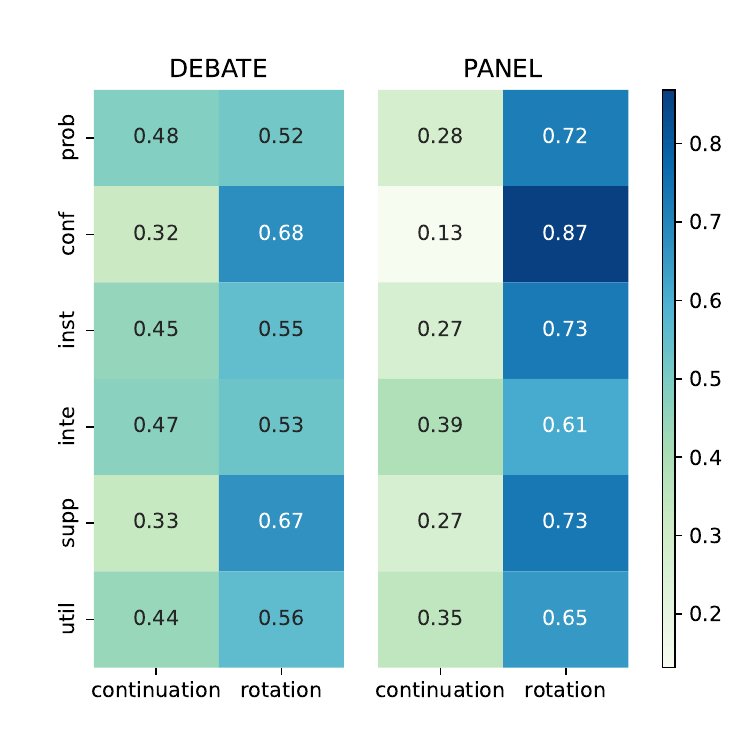}
\centering
\caption{Probabilities of participants' rotation statuses following different moderation dialogue acts.}
\label{fig:after_mod}
\end{figure}
We next use the dialogue act (DAs) to further investigate how rotations are facilitated. For each moderator turn $t$ (i.e.\ $p_t =$ moderator) and denoting the dialogue acts as $d_t$, we compute $P(s_{t+1}|d_t)$ and present the results in Figure \ref{fig:after_mod}.\footnote{Each moderator turn may have one or more dialogue acts (since there can be multiple sentences for a turn). All dialogue acts contribute to the transition matrix counts. For example, if $d_t = [\text{prob}, \text{inst}]$ and $s_{t+1} = rotation$, we have 2 transitions: $\text{prob} \rightarrow rotation$ and $\text{inst} \rightarrow rotation$.}
We see that moderator intervention  in \RTRP tends to lead to speaker \textit{rotation} across all dialogue acts. Most dialogue acts in \INSQ, however, lead to both \textit{continuation} and \textit{rotation} almost equally; the only exceptions are confronting and supplementary. This is perhaps not surprising, as confronting questions are designed to explicitly prompt one speaker to respond to another speaker's statement.

\subsection{Moderators' selection of target speakers}

\begin{table}
\centering
\resizebox{\columnwidth}{!}{%
\begin{tabular}{lccc} 
\toprule
 & {\bf Pro-activity} & {\bf Interactivity} & {\bf Specificity} \\ \midrule
\INSQ & 0.59 & 0.73 & 0.63 \\
\RTRP & 0.61 & 0.75 & 0.85 \\ \bottomrule
\end{tabular}%
}
\caption{Proportion of moderator sentences that are pro-active ($target\_speaker \neq last\_speaker$), interactive ($target\_speaker = next\_speaker$), and specific (targeting an individual).}
\label{tab:who}
\end{table}

Moderators may exhibit different interaction styles with participants depending on the context, in terms of pro-activity (how often the moderator actively initiates conversations), interactivity (how likely participants respond to the moderator), and specificity (how often the moderator addresses specific individuals rather than the group as a whole)~\citep{wagner2022comparing}. For instance, in a highly scripted setting, a moderator may act primarily as an assistant, responding to participant queries and broadcasting reminders about time and rules, with no expectation of responses—showing low levels of pro-activity, interactivity, and specificity. In contrast, in a more dynamic setting, a moderator might initiate conversations by asking questions tailored to individual speakers.

By analyzing whether the moderator’s target speaker aligns with the speakers preceding and following their intervention, we can infer the moderator's interaction style.
For each moderator turn $t$ (i.e.\ $p_t =$ moderator), we denote the set of target speakers as $r_t$ (a moderator turn can have multiple sentences and hence multiple target speakers) and compute pro-activity, interactivity and specificity as follows:
\begin{align*}
\text{pro-activity} &= \frac{\#(p_{t-1} \notin r_t)}{M} \\
\text{interactivity} &= \frac{\#(p_{t+1} \in r_t)}{M} \\
\text{specificity} &= \frac{\#(r_t \subset S)}{M}
\end{align*}
where $M$ is the total number of moderator turns and $S$ the set of unique participants in the conversation.

Table~\ref{tab:who} indicates that moderators in both domains demonstrate high levels of pro-activity and interactivity, suggesting that they frequently initiate interactions with participants. However, \RTRP moderators exhibit higher levels of specificity compared to \INSQ moderators, indicating a greater tendency to address specific individuals rather than the group as a whole. This suggests that \RTRP moderators are more likely to tailor their interventions to particular participants, fostering more targeted and personalized interactions.

\section{Conclusion}

We present WHoW, an analytical framework that characterizes conversational moderation across domains. WHoW breaks down the complexity of moderation decision-making into three key components: why the moderator intervenes (motives), how they intervene (dialogue acts), and to whom they direct the intervention (target speakers). Using this framework, we annotated moderation utterances in two distinct scenarios: Intelligent Squared Debate Corpus (\INSQ) and RoundTable Radio Panel Discussion (\RTRP). We showed that GPT-4o can effectively automate the labelling process. In total, our dataset has
5,657 human-annotated and 15,494 GPT-4o annotated moderation sentences, which is an order of magnitude larger than existing datasets \citep{park2012facilitative}.

Our analysis demonstrates the framework's effectiveness in differentiating intervention strategies and styles across the two scenarios. In \INSQ, moderators are primarily coordination-motivated, serving functional roles as interviewers and instructors, while occasionally facilitating interaction between non-moderator speakers. In contrast, \RTRP moderators are more information-oriented, acting as both contributors and interviewers, as they often participate in the discussion topics. While they seek information from the speakers and balance turn-taking, they promote less direct interaction between non-moderator speakers.

Our framework can serve as an exploratory tool or foundational skeleton for domain-specific adaptation and expansion in moderation analysis or moderator agent development. Using raw transcripts, users can initially categorize the moderator's speech into the twelve categories across the motive and dialogue act dimensions, then refine these labels based on the specific domain context. For example, in a mental health support setting, ``social interpretation'' could be expanded into more specific categories like ``emotion interpretation''. Although the current dataset may not be large enough for fine-tuning supervised models, it serves as a valuable resource for in-context few-shot learning. This means it provides practical examples that help develop models capable of predicting or recommending intuitive moderation interventions based on our framework.

Future studies should encompass a broader range of moderation scenarios, such as group counseling~\citep{kissil2016facilitators} and second language group conversations~\citep{gao2024interaction}. Additionally, the proposed analytic framework could be expanded to support the generation of conversational moderation strategies by sequentially predicting the three key components. Another important direction is the development of evaluation metrics to assess the effects and potential biases of moderation interventions~\citep{spada2013moderates}, enabling deeper insights into the impact and fairness of moderation practices. Finally, a broader goal for future work could involve synthesizing the results into ``moderator prototype strategies'' --- a schema with multiple axes capturing distinct moderation styles. As new scenarios are explored, these prototypes may evolve, offering a richer understanding of diverse moderation approaches.

\section{Limitations}

Some dimensions exhibit low to moderate inter-annotator agreement and low macro-F1 scores, indicating that the boundaries between certain concepts can be ambiguous and subjective. This issue is not unique to our research, as previous studies on moderation-related annotations have also reported both low~\citep{falk2024moderation} and high~\citep{park2012facilitative} levels of inter-annotator agreement. As shown in Table \ref{tab:agreement}, the agreement levels and macro-F1 scores differ across the settings we analyzed, suggesting that ambiguity is highly context-dependent, with some contexts using more explicit language and others relying on implicit expressions. We recommend that future studies adapting this framework incorporate some degree of human validation tailored to the specific context. Additionally, while we aimed to develop and validate an analytic framework that generalizes across scenarios, the two selected scenarios share a high degree of similarity, both placing less emphasis on social motives. This limitation was due to the lack of sufficient data to compare more diverse scenarios, as multi-party conversation data with clearly tagged moderators are scarce. However, despite the similarity between the selected scenarios, the framework successfully differentiated the two settings, demonstrating its potential for comparative analysis.

\section{Ethics Statement}

This study was conducted in accordance with the ACL Code of Ethics. Given that the multi-party discussion transcripts may involve controversial topics, annotators were informed in advance and were granted the right to skip any content they found uncomfortable. All identifiable personal information of the annotators has been removed from the datasets. Since the annotations are based on publicly available datasets ~\citep{zhang2016conversational, majumder2020interview}, there are no confidentiality concerns regarding the speakers' privacy or personal information. The annotation protocol and material were approved by the University of Melbourne research ethics committee with the reference code- 2023-28400-47354-1.

In terms of potential risks and dangers, our work at this stage is primarily analytical and does not involve content generation, thereby minimizing the risk of producing harmful material. Additionally, since the research focuses on moderation rather than persuasion, the findings are unlikely to contribute to harmful uses, such as the spread of propaganda.

\section*{Acknowledgments}

I would like to express my gratitude to Rena Gao, Rui Xin, Gisela Vallejo, and Miao Li, my wonderful colleagues who contributed significantly during the pilot study stage. I also wish to thank the anonymous workers who participated in the annotation process for this project. Additionally, I am grateful to Alex Goddard for sharing inspiring literature and insights that greatly informed this research.

\bibliography{custom}

\onecolumn
\appendix

\section{Appendix: Framework Supplementary Information}

\begin{table}[H]
\centering
\resizebox{\columnwidth}{!}{%
\begin{tabular}{|p{0.05\linewidth} | p{0.35\linewidth} p{0.27\linewidth} p{0.27\linewidth} | p{0.25\linewidth}|}
\toprule
DAs & IM & SM & CM & Source No. \\ \midrule
\rowcolor[HTML]{EFEFEF} Prob & asking users to provide more infomratino (0), asking user to make or consider possible solution (0), Posing a question at large for the users to respond(0), asking questions (1), asking for elaboration (1), asking for clarification and explanation (1), facilitating students’ argumentation (2), conversation stimulator (3), invite feedback or comments(5), questioning clarifications probe viewpoints(5), open invitation (6), specific intivation to participate (6), follow-up question (6) & empathetic exploration(4), participation encouragement (7) & coordinative enquiry* & \cellcolor[HTML]{FFFFFF}0: \citet{park2012facilitative},\newline1: \citet{vasodavan2020moderation},\newline2: \citet{hsieh2012effect},\newline3: \citet{wright2009role},\newline4: \citet{sharma2020computational},\newline5: \citet{lim2011critical},\newline6: \citet{schroeder2024fora},\newline7: \citet{mao2024multi},\newline*: observed from \citet{zhang2016conversational} \\
\rowcolor[HTML]{FFFFFF} Conf & encourage users to consider/engage comments of others (0), playing devil's advocate (1), helping students to sustain threaded discussion (2), problem solver (3), make connections (6) & conflict resolver (3), conflict resolution (7) & coordinative consensus building* & \cellcolor[HTML]{FFFFFF} \\
\rowcolor[HTML]{EFEFEF} Inst & Indicating irrelevant, offpoint comments (0), promote self-regulation (1), helping students focus on the main topics (2) & invite for team collaboration (1), & directing user to another more relevent issue post more relevent(0), redact and quarantine for inappropriate language content(0), maintaining/encouraging civil deliberative discourse(0), coordinating and planning (1), open censor (3), covert censor (3), cleaner (3), establish new threads/directions (5) & \cellcolor[HTML]{FFFFFF} \\
\rowcolor[HTML]{FFFFFF} Inte & correcting misstatements or clarifying (0), summarizaing discussion (1), highlight contribution (1), archiving information (1), summarizer of debates (3), summarize salient points (5), initiative summarization (7) & empathetic interpretation(4) & preference intepretation* & \cellcolor[HTML]{FFFFFF} \\
\rowcolor[HTML]{EFEFEF} Supp & providing information about the proposed rule (0), pointing to relevant information(0), pointing out characteristics of effective commenting(0), providing opinion (1), giving feedback (1), introduce other relevant information (1), providing judgment (1), constructive feedback (1), self evaluation (1), giving students positive feedback (2), supporter (3), ‘ ybrarian’ (3), expressing agreements(5), challenge others’ viewpoints (5), make connections with supporting research (5), providing opinions/explanation (5), express agreements or affirmation (6), model examples (6), in-context chime-in (7) & informal talk (1), adding personal experience/opinion (1), welcomer (3), empathetic reaction(4), direct chatting (7) & explaining the goals/rules of moderation(0), explaining the role of CeRI(0), explaining why comment is outside scope (0), & \cellcolor[HTML]{FFFFFF} \\
\rowcolor[HTML]{FFFFFF} Util & acknowledgement* & greeting (1), appreciation (1), humor (1), use emojis (1), making people feel welcome(3), acknowledgement or showing appreciation (5), express appreciation (6) & keep silent (7), floor grabbing* & \cellcolor[HTML]{FFFFFF} \\ \bottomrule
\end{tabular}%
}
\caption{This table presents a collection of literature with taxonomies for moderation/facilitation, mapping their classifications across the dialogue acts and motives dimensions of our framework.}
\label{tab:gen}
\end{table}

\begin{table}[H]
\centering
\resizebox{\columnwidth}{!}{%
\begin{tabular}{|p{0.05\linewidth} | p{0.35\linewidth} p{0.35\linewidth} p{0.35\linewidth}|}
\toprule
DAs & IM & CM & SM \\ \midrule
\rowcolor[HTML]{EFEFEF} 
Prob & Can you take that on? (prompting)\newline As long as the political spectrum is covered overall, what’s wrong with that? (follow up question)\newline Siva? (name   calling prompt) & Which of you would like to go first? (preference inquiry)\newline Did this gentleman come down yet? (coordinative question)\newline It's working, right? (question managing environment) & Is that a relief to you or-- (asking feeling)\newline Could you tell us your name, please? (social question)\newline Do you have eyeglasses? (humour question) \\ \midrule
Conf & That landed pretty well I think, so can you respond to that? (counter confronting)\newline On this side, do you want to respond, or do you agree? (consensus confronting)\newline You actually asked a perfect question, and so Mark Zandi, do you want to take that on? (confronting question) & The other side care to respond, if not I’ll move on.(coordinative consensus)\newline Response from the other side, or do you want to pass? (coordinative confronting)\newline Marc Thiessen, do you want to join your partner on this one, because I think-- (coordinative consensus) & Bryan Caplan, I think he just described your fantasy, come true.(social confronting)\newline I'd love to hear your answer to that question, so go for it. (confronting with affective appeal)\newline Jared Bernstein, the guy you called “nuts” just said you're unfair. (humour confronting) \\ \midrule
\rowcolor[HTML]{EFEFEF} 
Inst & Can you frame your question as a question? (articulate instruction)\newline Relate that point to this motion. (back to topic)\newline I want to stay on the   merits of the Obama plan. (manage topic) & Remember, about 30 seconds is what you'll get. (time control)\newline Can you go up three steps, please, and turn right? (coordinating instruction)\newline I'll be right back after this message. (program management) & Do not be afraid. (emotion instruction)\newline Those who agree, just a round of applause to that. (pro-social instruction)\newline --because it's turning into a personal attack. (stop anti-social) \\ \midrule
Inte & So, Matt, you're saying that it's not true that it's inevitable that Amazon will control everything. (summarization)\newline Their point is that it would be a bad thing. (simplification)\newline But that would be the question of mobility. (reframe) & That was an ambiguous signal. (situation interpretation)\newline You're pointing to Lawrence Korb.(preference interpretation)\newline And you want the side arguing   for the motion to address that (preference interpretation) & I think it was a rhetorical question, and it got a good laugh. (humour interpretation)\newline And it's a little bit insulting almost to say (toxicity interpretation)\newline —honestly, I don’t think that was an—a personal attack— (toxicity interpretation) \\ \midrule
\rowcolor[HTML]{EFEFEF} 
Supp & I agree that it is.(agreement)\newline The fact is that one of the US manufacturers, with 1 percent of its yearly production, would run us out of the whole market.(add information)\newline They had never paid any attention whatsoever to Africa. (share opinion) & Fifty-one of you voted against the motion. (vote reporting)\newline And the mic’s coming down to you. (describe situation)\newline Round two is where the debaters   address each other directly (rule explanation) & You have a colorful sleeve. (social chit-chat)\newline I hate to reward it but I'm going to. (encouragement)\newline And I think all of us probably share a sense that we want things to improve. (state common feeling) \\ \midrule
Util & Fair question. (acknowledgement)\newline Right (acknowledgement)\newline  So the-- (floor grabbing) & All right. (backchanneling)\newline Actually, I-- (floor grabbing)\newline Well—(floor   grabbing) & Thank you Evgeny Morozov. (thanks)\newline I'm sorry. (apology)\newline Hi. (greeting) \\ \bottomrule
\end{tabular}%
}
\caption{This table presents a collection of exemplar sentences at the intersection of the motives and dialogue acts dimensions.}
\label{tab:example_sents}
\end{table}

\begin{figure}[H]
\includegraphics[width=\columnwidth]{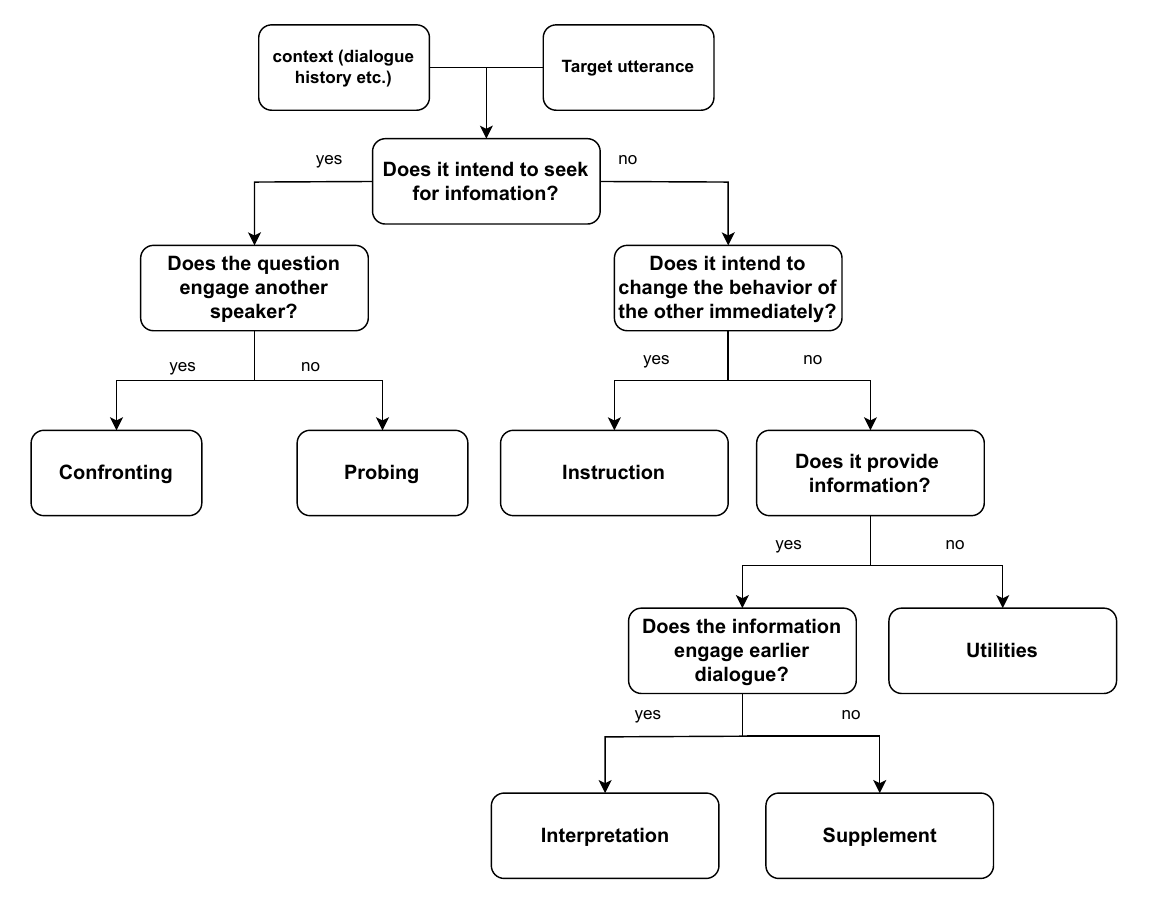}
\centering
\caption{The decision tree used by annotators to resolve ambiguous sentences that may involve multiple dialogue acts.}
\label{fig:da_tree}
\end{figure}

\section{Prompt Engineering}
\label{sec:prompt}

\begin{figure}[H]
\includegraphics[width=\columnwidth]{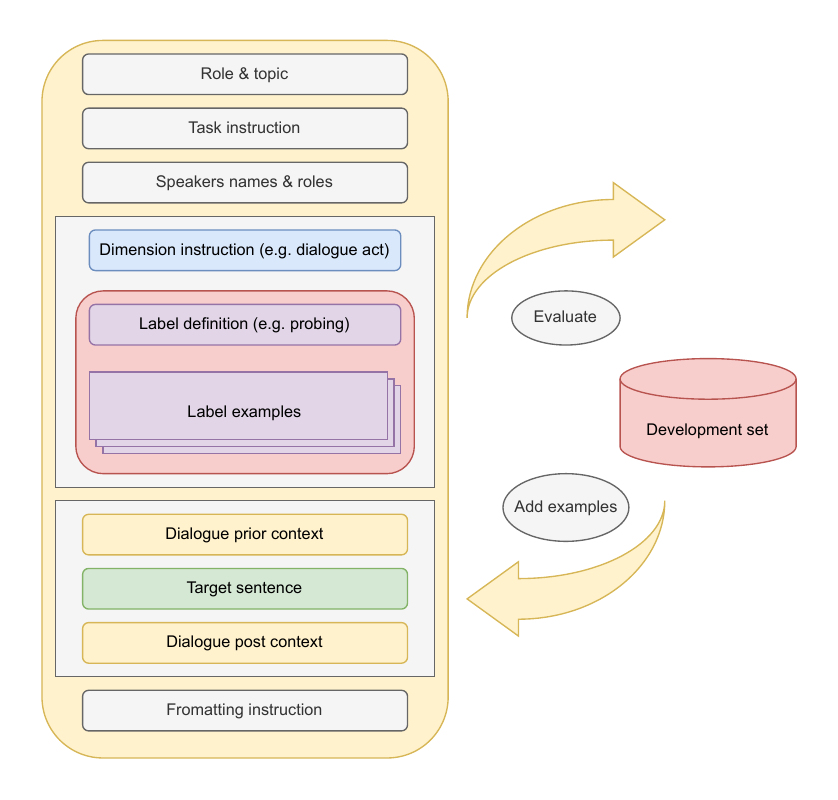}
\centering
\caption{Prompt structure and development cycle}
\label{fig:prompt}
\end{figure}

\begin{table}[]
\resizebox{\textwidth}{!}{%
\begin{tabular}{|p{0.170\linewidth} | p{\linewidth}|}
\toprule
section & prompt part \\ \midrule
\rowcolor[HTML]{EFEFEF} 
Role \& topic & Your role is an annotator, annotating the moderation behavior and speech of a debate TV show. The debate topic is ``When It Comes To Politics, The Internet Is Closing Our Minds" \\
Task instruction & given the definition and the examples, the context of prior and posterior dialogue, please label if the target utterance carries informational motive? \\
\rowcolor[HTML]{EFEFEF} 
Dimension instruction & Motives: During the dialogue, the moderator is acting upon a mixed-motives scenario, where different motives are expressed through responses depending on the context of the dialogue. Motives are the high level motivation that the moderator aim to achieve. The definitions and examples of the informational motive are below: \\
Label definition & informational motive: Provide or acquire relevant information to constructively advance the topic or goal of the conversation. \\
\rowcolor[HTML]{EFEFEF} 
Label examples & examples: “Why do you think minimum wage is unfair?” (Relevant information seeking.) “The legal system has many loopholes.” (Expressing opinion.) “Yea! I agree with your point!” (Agreement relevant to the topic.)  “The law was established in 1998.” (Providing topic relevant information.) \\
Dialogue prior context & Dialogue context before the target sentence:\newline\newline Eli Pariser (for): Just a little story, when I was on the book tour for my book, I was on a radio show in St. Louis. And the host decided to make this big spectacle of having people Google Barack Obama and call-in and read their search results. It was a really boring radio hour. And the first person called in, the second person called in and they interviewed everybody and had people kind of do a read-off where they’re both reading it off at the same time and it was exactly the same. And I was thinking, this is the worse book promotion I’ve ever done. And then a third guy called in, and he said you know it’s the damndest thing, when I Google Barack Obama, the first thing that comes up is this link to this site about how he’s not a natural citizen. And the second link is also a link to a website about how he doesn’t have a birth certificate.\newline\newline  Evgeny Morozov (against): That was your publicist.\newline\newline  Eli Pariser (for): Oh, I was wondering about that. But so, I think the danger here is that it’s not just that he was getting a view of the world that was really far off the average here. But he didn’t even know that that was the view that he was getting. He had no idea how tilted that view was. And that’s sort of the challenge. I just want to address one other point, which is that there seems to be this question about whether this is happening. And it’s really kind of funny to me, because if you talk to these companies and if you listen to what they’re saying, all of these companies are very clear that personalization is a big part of what they’re doing and what they're--\newline\newline  Evgeny Morozov (against): For pizza, weighted decisions. They are very clear. And they say we don’t want to do it for politics, we only want to do it for pizza.\newline\newline  Eli Pariser (for): Right, and the question is, can you trust them?\newline\newline  John Donvan (mod): Let me-- Jacob, I think Eli left a pretty good image hanging out there, of these folks truly not knowing how much they don’t know and believing what they’re getting and not understanding how slanted it is. \\
\rowcolor[HTML]{EFEFEF} 
Target sentence & Target sentence:\newline\newline John Donvan (mod): That landed pretty well I think, so can you respond to that? \\
Dialogue post context & Dialogue context after the target sentence:\newline\newline Jacob Weisberg (against): But a guy who called into a radio show? I know the plural of anecdote is data. But I mean, if this were really happening in the way you say it is, wouldn’t there be some kind of decent study that actually showed widely varying results? I mean as I say, I’ve tried to test this out as best I can. I’ve tried it myself on various browsers, signed in, signed out, Wikipedia always comes up first, sometimes it comes up second. Wikipedia’s vaccine entry is pretty good. I do not think there is actually the kind of variety you’re talking in searches done most of the time by most people.\newline\newline  John Donvan (mod): Siva. \\
\rowcolor[HTML]{EFEFEF} 
Formatting instruction & Please answer only for the target sentence with the JSON format:\{"verdict": 0 or 1,"reason": String\} For example:  answer: \{"verdict": 1, "reason": "The moderator asks a question to Joe Smith aimed at eliciting his viewpoint or reaction to a statement from the recent policy change for combatting climate change......"\} \\ \bottomrule
\end{tabular}%
}
\caption{An example of a single task prompt to determine if the target sentence has informational motive.}
\label{tab:single_prompt}
\end{table}

\begin{table}[]
\resizebox{\textwidth}{!}{%
\begin{tabular}{|p{0.15\linewidth} | p{1.3\linewidth}|}
\toprule
section & prompt part \\ \midrule
\rowcolor[HTML]{EFEFEF} 
Task instruction & given the definition and the examples, the context of prior and posterior dialogue, please label which motives the target response carries? And which dialogue act the target sentence belong to? And who is the moderator talking to? \\
Motives section & Motives: During the dialogue, the moderator is acting upon a mixed-motives scenario, where different motives are expressed through responses depending on the context of the dialogue. Different from dialogue act, motives are the high level motivation that the moderator aim to achieve. The definitions and examples of the 3 motives are below:\newline\newline informational motive: Provide or acquire relevant information to constructively advance the topic or goal of the conversation.  examples: “Why do you think minimum wage is unfair?” (Relevant information seeking.) “The legal system has many loopholes.” (Expressing opinion.) “Yea! I agree with your point!” (Agreement relevant to the topic.)  “The law was established in 1998.” (Providing topic relevant information.)\newline\newline  social motive: Enhance the social atmosphere and connections among participants by addressing feelings, emotions, and interpersonal dynamics within the group. examples: “It is sad to hear the news of the tragedy.” (Expressing emotion and feeling.) “Thank you! Mr. Wang.” (Appreciating.) “Hello! Let’s welcome Dr. Frankton.” (Greeting.) “I can understand your struggle being a single mum.” (Empathy) “How do you feel? when your work was totally denied.” (Exploring other’s feeling.) “Please feel free to say your mind because I can’t bite you online, hehe!” (Humour.) “The definition is short and simple! I love it!” (Encouragement.) “Maybe Amy’s intention is different to what you thought, you guys actually believe the same thing.” (Social Reframing.)\newline\newline  coordinative motive: Ensure adherence to rules, plans, and broader contextual constraints, such as time and environment. examples: “Let’s move on to the next question due to time running out.” (Command) “We going to start with the blue team and then the red team” (Planning) “Do you want to go first?” (Asking for process preference.) “Please move to the left side and turn on your mic!” (Managing environment) \\
\rowcolor[HTML]{EFEFEF} 
Dialogue act section & Dialogue act: Dialogue acts is referring to the function of a piece of a speech. The definitions and examples of the 6 motives are below:\newline\newline Probing: Prompt speaker for responses. examples: “What is your view on that Dr. Foster?” (Questioning.) “Where are you from?” (Social questioning.) “Peter!” (Name calling for response.) “If the majority of people are voting against it, would you still insist?” (Elaborated questioning.) “Do you agree with this statement?” (Binary question.)\newline\newline Confronting: Prompt one speaker to response or engage with another speaker's statement, question or opinion. examples: “So David pointed out the critical weakness of the system, what is your thought on his critiques, Dr. Foster?”, "Judge Anderson, what is your response to this hypothetical scenario posed by Ms. Lee regarding privacy laws?", "Senator Harris, you have proposed reducing taxes instead. How do you respond to Mr. Walkers suggestion to increase school funding?", "So, Dr. Green, Professor Brown just criticized the emissions policy. What is your response to his critique?"\newline\newline Supplement: Enrich the conversation by supplementing  details or information without immediately changing the target speaker's behavior. examples: “And that concludes round one of this Intelligence Squared U.S. debate where our motion is Break up the Big Banks.” (Addressing progess) “The blue team will go first, then the red team can speak” (explaining program rule) “Supposed we live in a world where such behaviour is accepted.” (Hypothesis) “I suggest the best solution is giving everyone equal chances.” (Proposal) “The government announced tax raise from March.” (Providing external information) “I agree with that you said.” (Agreement) “GM means genetic modified.” (Providing external knowledge) “I think people should be given the right to say no!” (Opinion) "The guy with the blue shirt." (Describing appearance) "The power is off." (Describing situation). “In this section, debaters will address one another and also take questions from the audience.” (Explaining upcoming segment) "Let me move this along a little bit further to a slightly different topic, although we have circled around it." (Explaining self intention) "I want to remind you that we are in the question and answer section." (Remind current phase of the discussion)\newline\newline  Interpretation: Clarify, reframe, summarize, paraphrase, or make connection to earlier conversation content. examples: “So basically, what Amy said is that they didn’t use the budget efficiently”. (Summarization) “You said ‘I believe GM is harmless,’.” (Quote) “In another word, you don’t like their plan.”. (Paraphrase) “My understanding is you don’t support this due to moral reason.” (Interpretation) “She does not mean to hurt you but just tell the truth.” (Clarify) “So far, we have Dr. Johnson suggesting…., and Dr. Brown against it because……”(Summarization) “Amy saying that to justify the reduction of the wage, but not aiming to induce suffering.” (Reframing)\newline\newline  Instruction: Explicitly command, influence, halt, or shape the immediate behavior of the recipients. examples: “Please get back to the topic.” (Commanding) “Please stop here, we are running out of time.” (Reminding of the rule) “The red will start now.” (Instruction) “Please mind your choice of words and manner.” (social policing) “Do not intentionally create misconception.” (argumentative policing) “Now is not your term, stop here.” (coordinative policing) “What you need to do is raise your hand, and ushers will come to you.” (Guiding participation) “Turn on your microphone before speaking.” (Technical instruction)  All Utility: All other unspecified acts.  examples: “Thanks, you.” (Greeting) “Sorry.” (Apology) “Okay.” (Back channelling) “Um hm.” (Back channelling) “But, but, but……” (Floor grabbing) \\ 
Formatting instruction & Please answer only for the target sentence with the JSON format:\{"motives": List(None or more from "informational motive", "social motive", "coordinative motive"),"dialogue act": String(one option from "Probing", "Confronting", "Supplement", "Interpretation", "Instruction", "All Utility"),"target speaker(s)": String(one option from "0 (Unknown)", "1 (Self)", "2 (Everyone)", "3 (Audience)", "4 (Eli Pariser- for)", "5 (Siva Vaidhyanathan- for)", "6 (Evgeny Morozov- against)", "7 (Jacob Weisberg- against)", "8 (Support team)", "9 (Against team)", "10 (All speakers)"),"reason": String\}\newline\newline For example: answer: \{"motive": {[}"informational motive"{]}, "dialogue act": "Probing",  "target speaker(s)": "7 (Joe Smith- for)", "reason": "The moderator asks a question to Joe Smith aimed at eliciting his viewpoint or reaction to a statement from the recent policy change for combatting climate change......"\} \\ \bottomrule
\end{tabular}%
}
\caption{An example of a multi-task prompt. Here we only demonstrate the components that are different from the single-task prompt.}
\label{tab:multitask_prompt}
\end{table}

Our prompt design, as illustrated in Figure \ref{fig:prompt}, incorporates several key components: a concise description of the moderation scenario and the annotator’s role, an introduction to the task, an explanation of the dimensions and corresponding labels, five preceding responses for context, the target sentence, two subsequent responses for additional context, and instructions for the output format. The label instructions include both definitions from the annotation manual and single-sentence examples. We initially began with a few seed examples for each label and iteratively introduced new examples that had been misclassified during the development process to enhance performance. Table \ref{tab:single_prompt} provides a detailed example of a single-task prompt. Additionally, we developed a multi-task prompt that stacks all label definitions and examples across the three dimensions, with adjusted formatting instructions. Table \ref{tab:multitask_prompt} highlights the modifications and stacked elements of the prompt.

\subsection{Supervised model training and comparison}
\label{sec:longformer}

\begin{table}[H]
\setlength\tabcolsep{1.5pt}
\centering
\begin{adjustbox}{max width=\linewidth }
\begin{tabular}{lccccc}
\toprule
\textbf{Model} & \textbf{DA} & \textbf{IM} & \textbf{CM} & \textbf{SM} & \textbf{TS} \\
\midrule
Random (\INSQ)        & 0.153          & 0.492                   & 0.508            & 0.405                  & 0.057             \\
GPT-4o-MT(\INSQ)                 & 0.485          & 0.761                  & 0.711          & 0.767                 & 0.497             \\
GPT-4o-ST(\INSQ)     & \textbf{0.515}          & 0.729                  & 0.686           & 0.668                 & \textbf{0.525}            \\
longformer-MT(\INSQ)  & 0.494          & 0.764                 & 0.719           & \textbf{0.784}                & 0.246             \\
longformer-ST(\INSQ) & 0.493         & \textbf{0.772}                  & 0.726           & 0.694                 & 0.299             \\
DialogLMLED-MT(\INSQ) & 0.489         & 0.760                  & \textbf{0.760}           & 0.714                 & 0.147             \\
\midrule
Random(\RTRP)        & 0.115          & 0.490                   & 0.482            & 0.387                  & 0.096 \\
GPT-4o-MT(\RTRP)                 & \textbf{0.504}          & 0.726                  & 0.732           & 0.754                 & \textbf{0.467}             \\
GPT-4o-ST(\RTRP)     & 0.492          & 0.747                  & 0.639           & 0.635                 & 0.464            \\
longformer-MT(\RTRP)  & 0.414          & 0.753                 & \textbf{0.774}           & 0.731                & 0.196             \\
longformer-ST(\RTRP) & 0.417         & 0.757                 & 0.759          & 0.729                 & 0.225             \\
DialogLMLED-MT(\RTRP) & 0.389         & \textbf{0.764}                  & 0.751          & \textbf{0.768}                 & 0.132             \\
\bottomrule
\end{tabular}
\end{adjustbox}
\caption{Macro-F1 comparing GPT-4o and Longformer using multi-task (MT) and single-task (ST) approaches across the two subsets. The bold numbers highlights the top performer of the dimension in the subset. The random baseline is derived from five random simulations.}
\label{tab:longformer}
\end{table}

To further explore training smaller language models for motive and dialogue act classification, we fine-tuned the Hugging Face pre-trained Longformer model (allenai/longformer-base-4096)~\citep{beltagy2020longformer}. The input sequence included the discussion topic; a list of speaker options comprising all speaker names along with "unknown," "everyone," "audience," and "all speakers"; and, for the \INSQ\ subset, additional options "against team" and "support team." We also incorporated the five utterances preceding and the two utterances following the target sentence, with a maximum input length of 3,072 tokens. The model was trained for three epochs over three hours using a learning rate of 2e-5 with the AdamW optimizer (weight decay = 0.01) and a batch size of 8 on an A100 GPU via the Spartan cluster.

We compared both single-task and multi-task variants of the Longformer, employing individual and combined loss functions, respectively. For the multi-task approach, we adapted the model to include multiple classifier heads, each corresponding to a different classification task, and backpropagated using a combined loss function. Additionally, recognizing that the original Longformer models were not pre-trained on dialogue data, we included DialogueLM LED~\citep{zhong2022dialoglm}— a variant of Longformer model with a 5,120-token input context length and was pre-trained on interview and radio conversation corpora—in our experiments.

The results measured against the human-labeled test set are presented in Table~\ref{tab:longformer}. While the fine-tuned Longformer models demonstrated performance comparable to GPT-4o across most dimensions, they showed a notable disparity in predicting the target speaker. This discrepancy may be attributed to the dynamic nature of classification labels—the number and identity of speakers change between episodes.

Generative or retrieval approaches are more effective for target speaker classification. Finally, we observed that pre-training the model with dialogue corpora did not noticeably impact performance.

\section{Disagreement Cases Analysis}
\label{sec:disagree}

\begin{table}[H]
\resizebox{\textwidth}{!}{%
\begin{tabular}{{|p{0.2\linewidth} |p{0.85\linewidth}|}}
\toprule
Dimensions & Examples \\ \midrule
Dialogue act & 1. You know, what do you think about that, Callie? (prob vs. conf)\newline 2. Our time has run out. (supp vs. inst)\newline 3. Well let me move on to our final topic, which is gentrification. (supp vs. inst)\newline 4. Rick MacArthur cited Mexico, it has worked for Mexico.(supp vs. inte)\newline 5. Yeah. (supp vs. util) \\ \midrule
Motives & 6. Can you take that on? (IM vs. CM)\newline 7. Okay, go ahead. (IM vs. CM)\newline 8. Let's let Jacob Weisberg (IM vs. CM)\newline 9. So Lenny took the initiative of sending a question into us by email. (IM vs. SM)\newline 10. Do you agree that our nation needs affirmative action for intelligent conversation? (IM vs. SM)\newline 11. All right. (CM vs. SM) \\ \midrule
Target Speaker & 12. And that concludes round one of this Intelligence Squared US debate (everyone vs. audience)\newline 13. Let's bring Evgeny in and-- (everyone vs. Evgeny)\newline 14. And we also-- is Lenny Gengrinovich here? (everyone vs. Lenny) \\ \bottomrule
\end{tabular}%
}
\caption{Examples of disagreement cases across the dimensions of dialogue acts, motives, and target speaker. Bracketed information includes the combinations of disagreed labels. All examples are from the \INSQ dataset.}
\label{tab:disagree_examples}
\end{table}

In this appendix, we highlight the complexity and difficulty of the task by curating several examples in Table \ref{tab:disagree_examples}. We analyze and discuss cases of disagreement, particularly within the \INSQ subset, which received a relatively low agreement score. 

\begin{figure}[H]
\includegraphics[width=\columnwidth]{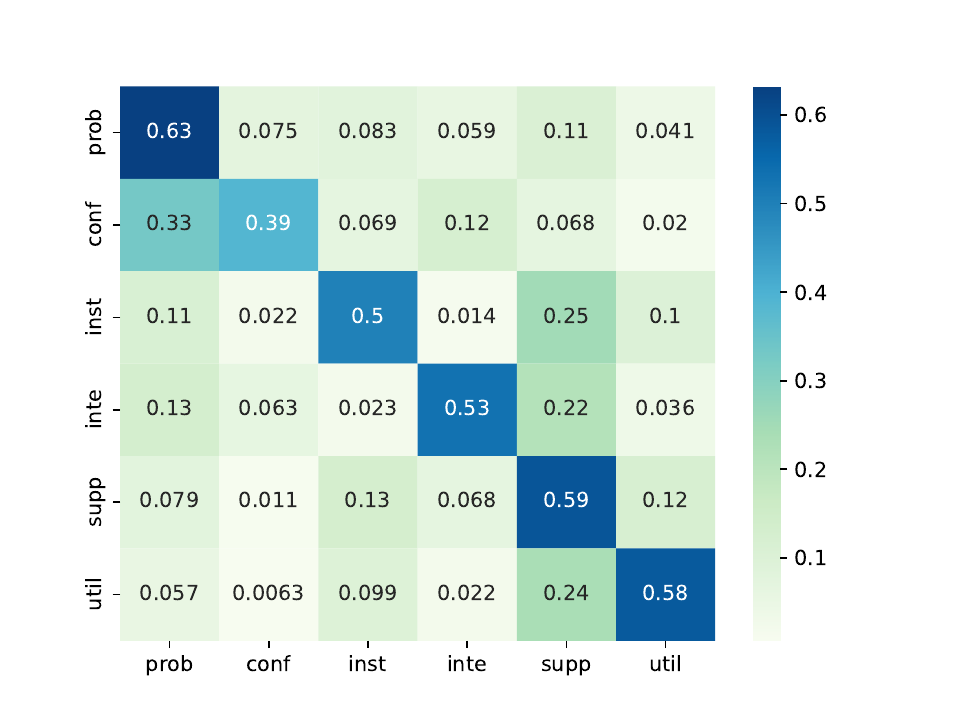}
\centering
\caption{The normalized co-occurrence matrix of dialogue act human votes from the \INSQ subset.}
\label{fig:da_disgree}
\end{figure}

To better understand the disagreements in dialogue act annotations, we calculated the co-occurrences of human annotators' votes, as shown in Figure \ref{fig:da_disgree}. While most dialogue act labels exhibit strong internal consistency, indicating general agreement among annotators, the figure reveals two primary sources of disagreement. The first source involves cases of 'confrontation,' where disagreement often arises when the moderator does not explicitly mention the intended participant by name, leading to differing interpretations of whether the confrontation is implied or direct (Example 1). The second source of disagreement involves the label 'supplement,' which frequently co-occurs with 'instruction,' 'interpretation,' and 'utility.' Examples 2 and 3 illustrate instances where it is unclear whether the moderator is expecting a behavioral change from the recipient or merely providing a reminder or explanation. Additionally, there are numerous ambiguous cases between 'supplement' and 'utility,' such as brief responses like 'Yeah,' where it is uncertain whether the expression is intended as acknowledgment or simple backchanneling.

For disagreements regarding motive labels, we found that the 'coordinative' motive was particularly often confused with the other two categories. Examples 6 to 8 highlight cases where vague probing led some annotators to interpret the moderator’s actions as rotating turns according to program rules, while others perceived the probing as an attempt to prompt information from the speakers to contribute to the topic. Short utility phrases like 'All right,' as seen in Example 10, also present ambiguity in motive—whether it’s meant for pacing or calming the speaker’s emotions is unclear. Additionally, disagreements were noted in the target speaker dimension. In Example 12, it is uncertain whether the moderator is addressing everyone or just the audience. Similarly, in Examples 13 and 14, the addressee shifts mid-sentence, leading to further confusion. 

These analyses underscore the inherent complexity and subjectivity involved in labeling dialogue acts and motives. Despite efforts to create clear definitions and guidelines, the nuanced nature of communication often results in differing interpretations among annotators, especially when dealing with implicit intentions, vague statements, or multi-functional phrases. 

\newpage

\section{Human Machine Annotation Comparative Analysis}
\label{sec:human_machine}
\begin{table}[H]
\centering
\begin{tabular}{@{}llllllll@{}}
\toprule
\multicolumn{8}{c}{\INSQ human}                                                     \\ \hline
      & prob           & conf  & inst           & inte  & supp           & util          & $p(m)$  \\ \hline
IM    & \textbf{0.48*} & 0.09  & 0.05           & \underline{0.25*} & 0.11           & 0.02          & 0.37   \\
CM    & 0.13           & 0.01  & \underline{0.27}           & 0.02  & \textbf{0.52*} & 0.04          & 0.53   \\
SM    & 0.04           & 0.01  & 0.03           & 0.01  & \underline{0.36*}          & \textbf{0.54}          & 0.12   \\ \hline
Total & 0.24*          & 0.04  & 0.15           & 0.09* & 0.34*          & 0.14          & \textit{150.77} \\ \hline
\multicolumn{8}{c}{\INSQ GPT-4o}                                                    \\ \hline
IM    & \textbf{0.40}  & \underline{0.22*} & 0.04           & 0.11  & \underline{0.22*}          & 0.01          & 0.39   \\
CM    & \underline{0.14}           & 0.10* & \textbf{0.54*} & 0.02  & 0.10           & 0.11*         & 0.66*  \\
SM    & 0.06           & 0.01  & 0.12*          & 0.02  & \underline{0.16}           & \textbf{0.64} & 0.12   \\ \hline
$p(d)$ & 0.22           & 0.11* & 0.36*          & 0.05  & 0.12           & 0.14          & \textit{150.77} \\ \hline
\multicolumn{8}{c}{\RTRP human}                                                     \\ \hline
      & prob           & conf  & inst           & inte  & supp           & util          & $p(m)$  \\ \hline
IM    & \textbf{0.51*} & 0.02  & 0.02           & 0.02  & \underline{0.42}           & 0.01          & 0.60   \\
CM    & 0.03           & 0.00  & \underline{0.09}           & 0.00  & \textbf{0.85*} & 0.03          & 0.28   \\
SM    & 0.00           & 0.00  & 0.01           & 0.02  & \underline{0.20}           & \textbf{0.72} & 0.06   \\ \hline
$p(d)$ & 0.31           & 0.01  & 0.03           & 0.02  & 0.55*          & 0.08          & \textit{61.35}  \\ \hline
\multicolumn{8}{c}{\RTRP GPT-4o}                                                    \\ \hline
IM    & \underline{0.41}           & 0.04* & 0.01           & 0.03  & \textbf{0.50*} & 0.01          & 0.72*  \\
CM    & 0.08*          & 0.02  & \textbf{0.41*} & 0.01  & \underline{0.33}           & 0.16*         & 0.25   \\
SM    & 0.05           & 0.00  & 0.02           & 0.01  & \underline{0.27}           & \textbf{0.64} & 0.16*  \\ \hline
$p(d)$ & 0.30           & 0.03* & 0.10*          & 0.02  & 0.41           & 0.13*         & \textit{61.35}  \\ \hline
\end{tabular}%
\caption{Conditional probabilities of dialogue acts (columns) given motives (rows), along with marginal probabilities of dialogue acts (right column) and motives (bottom row). All values are averaged across episodes from the \textbf{test} and \textbf{development} sets for the two scenarios—\INSQ (top) and \RTRP (bottom)—and for the two annotation sources: human and GPT-4o. The most frequent dialogue act for each motive is highlighted in bold, with the second most frequent underlined. The italicized number in the corner indicates the average frequency of moderator sentences. An * denotes values that are statistically significantly greater than their annotation source counterparts (human vs. GPT-4o; t-test at $p<=0.05$).}
\label{tab:human_machine}
\end{table}

To validate the analytical findings from the automatic system, we conducted a comparative study between human annotations and machine-generated annotations (using GPT-4o) on the test and development datasets. Table \ref{tab:human_machine} presents the conditional and marginal probabilities across the two settings (\INSQ vs. \RTRP) and the two annotation approaches (human vs. GPT-4o). Overall, the results indicate that machine annotations generally align well with human annotations. Although some differences are statistically significant, their magnitudes are typically small ($\leq 0.1$).

One notable exception is the distinction between coordinative-motivated "instruction" and "supplement." In our error analysis, we found that this discrepancy arises from differences in interpreting the "immediacy" of the expected influence on subsequent turns. An "instruction" act is intended for moderator interventions that expect an immediate change in the target speaker's behavior (e.g., "Please stay on topic."). In contrast, when moderators provide information without expecting immediate action (e.g., "After the debate, we will proceed to voting."), human annotators tend to label it as a "coordinative-motivated supplement," as it provides context or rules without requiring an immediate response. Machine annotations, however, did not consistently capture this nuance and often mislabeled these explanations of rules as "instructions," overlooking the subtle difference in immediacy.

We acknowledge the need to refine these aspects of the annotation framework to improve accuracy. Nevertheless, the core patterns and characteristics identified by both human and machine annotations remain largely consistent, reinforcing the validity of our primary findings.

\section{Classification Error Analysis}
\label{sec:error}

\begin{figure}[H]
\includegraphics[width=\textwidth]{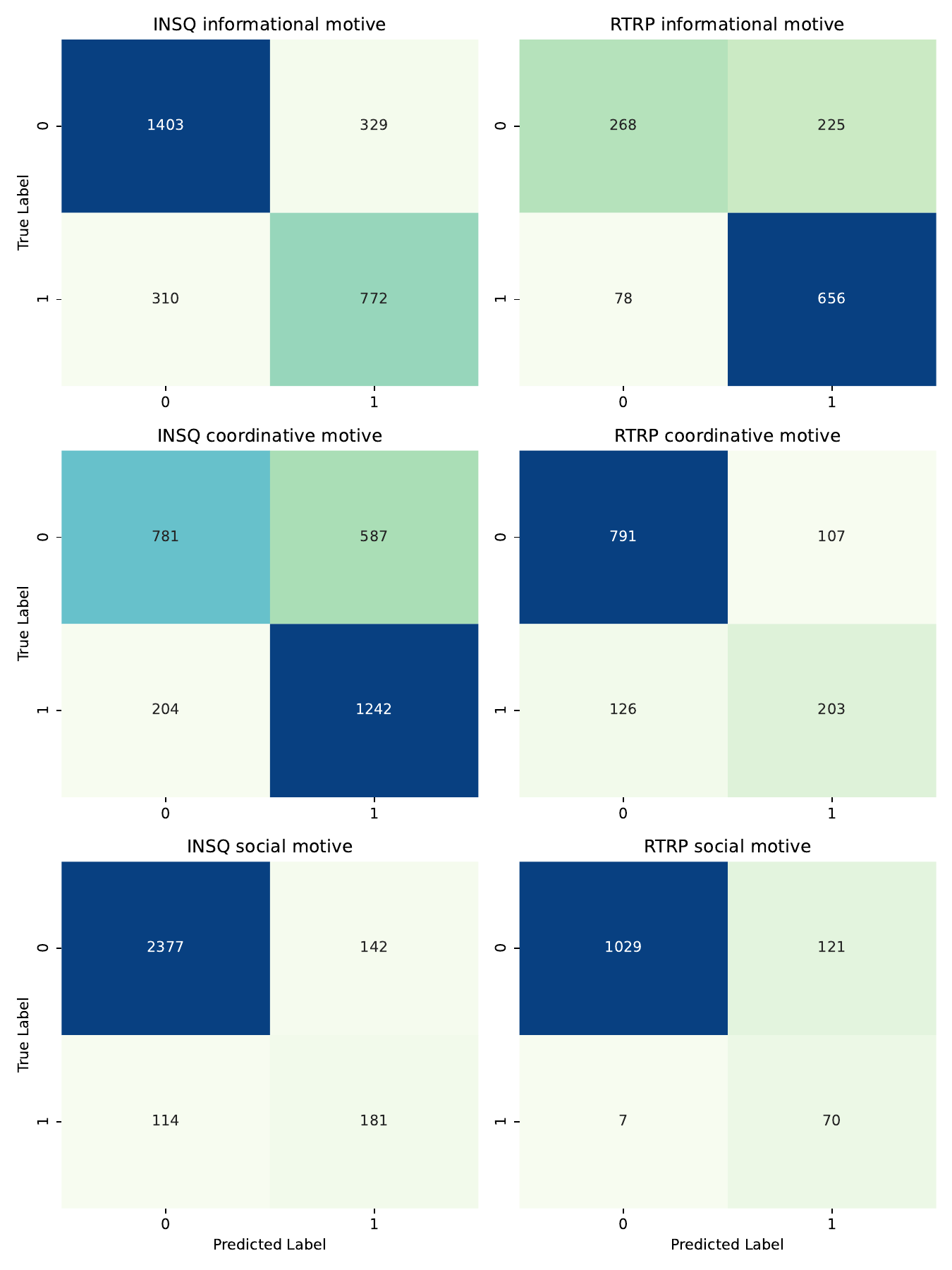}
\centering
\caption{The confusion matrices for the three motives across the two subsets.}
\label{fig:motive_consusion}
\end{figure}

\begin{figure}[H]
\includegraphics[width=0.8\textwidth]{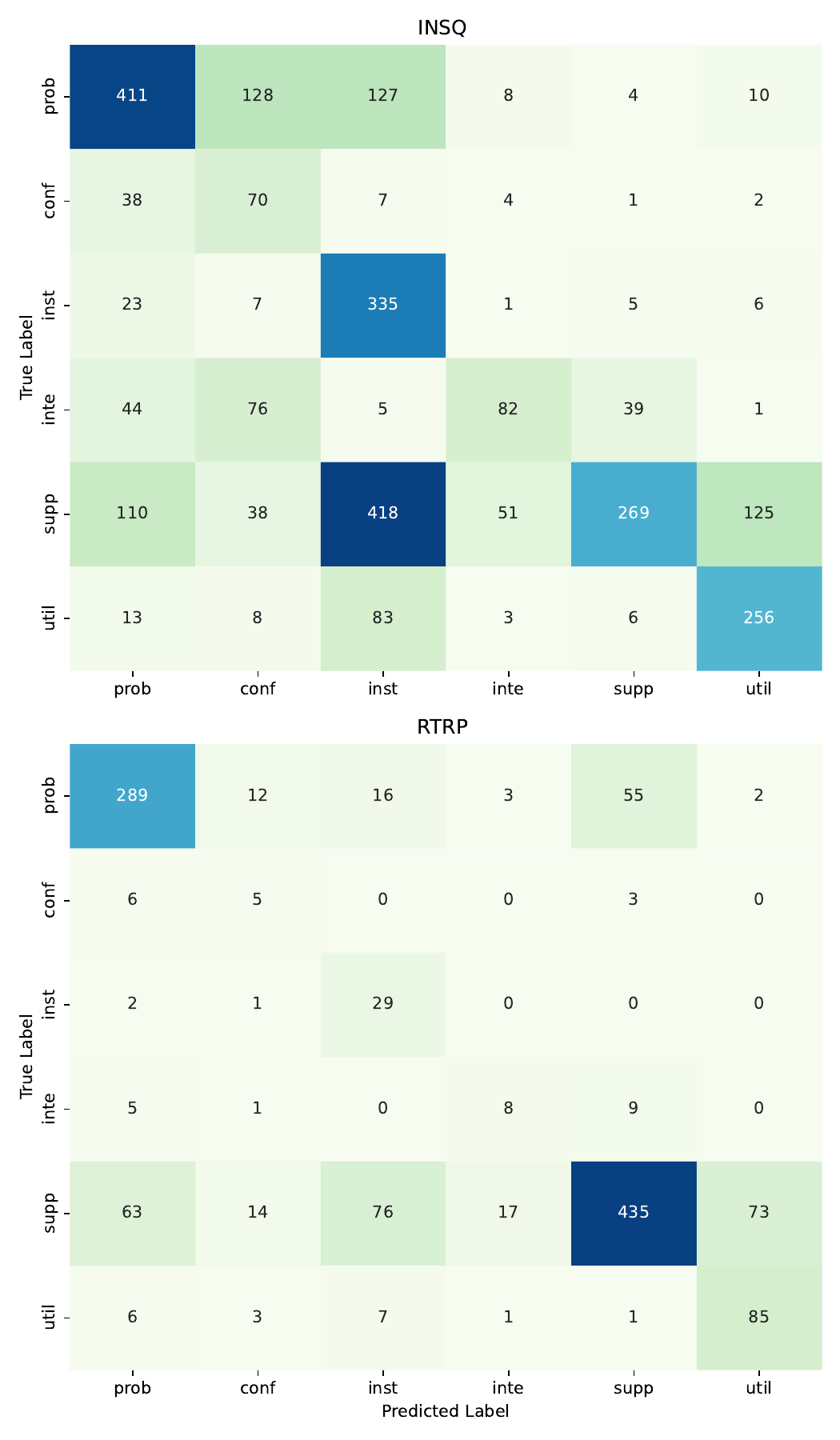}
\centering
\caption{The confusion matrices for the three motives across the two subsets.}
\label{fig:da_consusion}
\end{figure}

\begin{table}[H]
\resizebox{\textwidth}{!}{%
\begin{tabular}{{|p{0.15\linewidth} |p{\linewidth}|}}
\toprule
Dimensions & Examples \\ \midrule
Dialogue act & 1. Eli Pariser. (prob vs. conf, \INSQ).\newline
2. Dr. David Satcher. (conf vs. prob, \INSQ).\newline
3. I want to bring Matt back into this conversation. (prob vs. conf, \INSQ)\newline 
4. But wasn't your partner using the "that's what happened to me when I typed in Egypt"? (prob vs. inte, \INSQ)\newline
5. Let's go to Frank Foer. (prob vs. inst, \INSQ)\newline
6. There was a lot of questions that came up during Jena Six, saying, oh, marching is so 1965.(prob vs. supp, \RTRP)\newline
7. Your opponents are saying that Amazon cannot be trusted, that it's becoming more and more powerful, and that's probably likely to continue, although you're saying there are mitigating forces.(inte vs. conf, \INSQ)\newline
8. Also, that in a peace process that is going nowhere, that is stuck, it lays down a marker that the Israelis cannot ignore.' (inte vs. supp, \INSQ)\newline
9. I have a-- question in the second row. (supp vs. prob, \INSQ)\newline
10. You work for the Washington Post and I couldn't even find the story online about that. (supp vs. prob, \RTRP)\newline
11. We’re going to ask you to vote again at the end and the team that has moved its numbers the most will be declared our winner. (supp vs. inst, \INSQ)\newline
12. Microphones will be brought forward if you raise your hand. (supp vs. inst, RTRO)\newline
13. Yep (supp vs. util, \INSQ)\newline
14. Alright (util vs. inst, \INSQ) \\ \midrule
Motives & 15. So how would you relate that directly to the motion? (IM false positive, \INSQ)\newline
16. Jacob Weisberg. (IM false negative, \INSQ)\newline
17. What do you - Jasmyne, I'll start with you - unfold your, uncross your arms. (IM false negative, \RTRP)\newline
18. The team arguing against the motion, Franklin Foer and Scott Turow, they're saying, "It's all a trap. (CM false positive, \INSQ)\newline
19.Our motion is “America is to Blame for Mexico’s Drug War,” at the start, 43 percent of you were for…22 percent against, and 35 percent undecided. (CM false negative, \INSQ)\newline
20. Today on our Bloggers' Roundtable, we're taking a close look at urban education and the race for the White House. (CM false positive, \RTRP)\newline
21. Well, you're laughing because you think it's impossible or what is… (SM false positive, \RTRP)\newline
22. All right. (SM false negative, \INSQ) \\ \midrule
Target Speaker & 23. Round two is where the debaters address each other directly and also answer questions from you in the audience and from me. (audience vs. everyone, \INSQ)\newline
24. Let me ask the side that’s arguing that when it comes to politics, the internet is closing our minds. (support team vs. all speakers, \INSQ)\newline
25. But Evgeny kind of addressed that point when he-- I think you said, Evgeny, earlier in your opening statements, that initially the theory was the internet gave us tools to do stuff that we were already doing. (audience vs. Evgeny, \INSQ)\newline
26. Let me approach this from a couple of different angles. (all speakers vs. audience, \RTRP) \\ \bottomrule
\end{tabular}%
}
\caption{Examples of error cases across the dimensions of dialogue acts, motives, and target speaker. Bracketed information indicates the predicted labels vs. the human-aggregated labels, along with the source of each example.}
\label{tab:error_examples}
\end{table}

In this appendix, we examine the discrepancies between the GPT-4o-based classification results and the aggregated human annotation labels. Figure \ref{fig:motive_consusion} presents the confusion matrix for the three motives, comparing GPT-4o with the aggregated human annotations, while Figure \ref{fig:da_consusion} displays the confusion matrix for the six dialogue act labels. Table \ref{tab:error_examples} provides examples of common errors across the three dimensions to support further qualitative analysis.

An analysis of the dialogue act confusion matrix in Figure \ref{fig:da_consusion}, particularly within the \INSQ subset, reveals four primary sources of error. First, several probing sentences are frequently misclassified as confrontational or instructional. In Table \ref{tab:error_examples}, Examples 1 and 2 illustrate instances where the sentences merely include the addressees' names, and the intended purpose of the moderator—to engage the addressees with a previous speaker—depends heavily on the conversational context and remains inherently subjective. Ambiguous cases, such as Example 5, demonstrate scenarios where it is unclear whether the moderator is seeking information or simply inviting someone to participate. Additionally, long sentences may be reasonably associated with more than one dialogue act, as seen in Example 7, where both interpretation and confrontation are plausible classifications. A substantial number of errors also arise from confusion between 'supplement' and 'instruction,' which is the largest source of misclassifications. In Examples 11 and 12, it is often uncertain whether the moderator is merely explaining or reminding participants of a rule or the program’s progress, or if they expect a specific response. Lastly, numerous errors involve brief utility phrases like 'Yep' and 'Alright,' as in Examples 13 and 14. These phrases are highly context-dependent, making it challenging to determine whether the moderator is expressing acknowledgment, signaling the speaker to stop, or simply backchanneling.

Analyzing the confusion matrix for motive prediction in Figure \ref{fig:motive_consusion}, we identified two primary sources of error. In the \INSQ subset, the 'coordinative' motive exhibited the lowest performance, with most errors being false positives. For example, in Table \ref{tab:error_examples}, Example 18 involves the moderator introducing a key argument for the opposing team. Although this instance was annotated as driven by an informational motive, GPT-4o incorrectly interpretate it as an coordinative move for setting up the introduction. A similar pattern is observed in Example 20 from the \RTRP subset, where the moderator introduces the discussion’s background and topic. While GPT-4o classified this action as coordination-driven, human annotators labeled it as informational, despite one annotator also indicating a coordinative motive. Additionally, errors related to social motives proved particularly difficult to interpret, as seen in Examples 21 and 22.

In terms of target speaker classification errors, most misclassification occur when the target speaker is plural,e.g. "everyone". When multiple speakers are addressed, determining the scope or boundary of the intended recipients can be subjective and ambiguous. Examples 23, 24, and 26 illustrate the difficulty in discerning whether the moderator is addressing the entire group or only the audience. Another common source of error arises when the speaker shifts the intended recipient mid-sentence, as demonstrated in Example 25.

In our error case analysis, we identified several instances where GPT-4o classifications diverged from human annotations. However, these misalignments are not always unreasonable. Many examples are highly context-dependent, subjective, and open to interpretation, particularly in cases involving long sentences that could be associated with multiple labels or extremely short sentences, such as name-calling or backchanneling, where interpretation relies heavily on the conversational context. We also examined the reasons generated by GPT-4o to justify its classifications and found that, while they differ from the aggregated human annotations, the majority of these justifications are still defensible.

\section{Annotator instruction and material}
\label{sec:anno}

\begin{figure}[H]
\includegraphics[width=\textwidth]{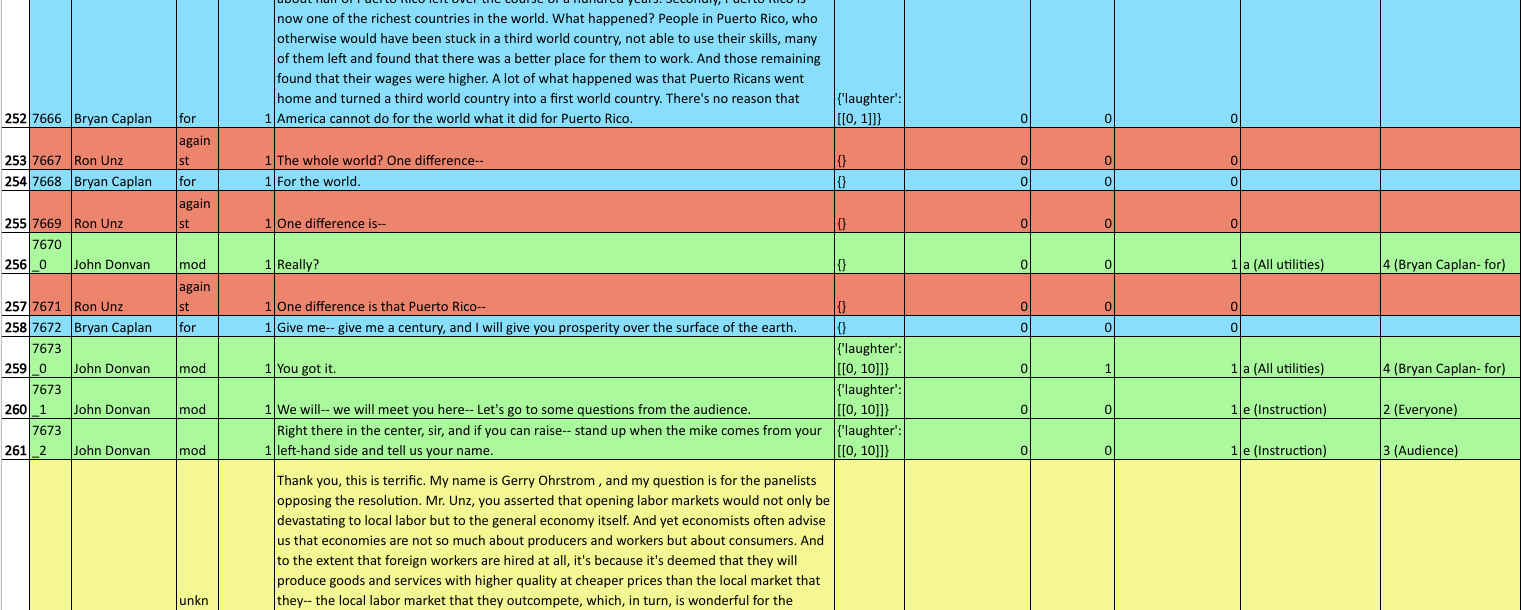}
\centering
\caption{The Excel sheet annotation interface used for annotating moderator transcript.} 
\label{fig:anno_inter}
\end{figure}

\includepdf[pages=1-10]{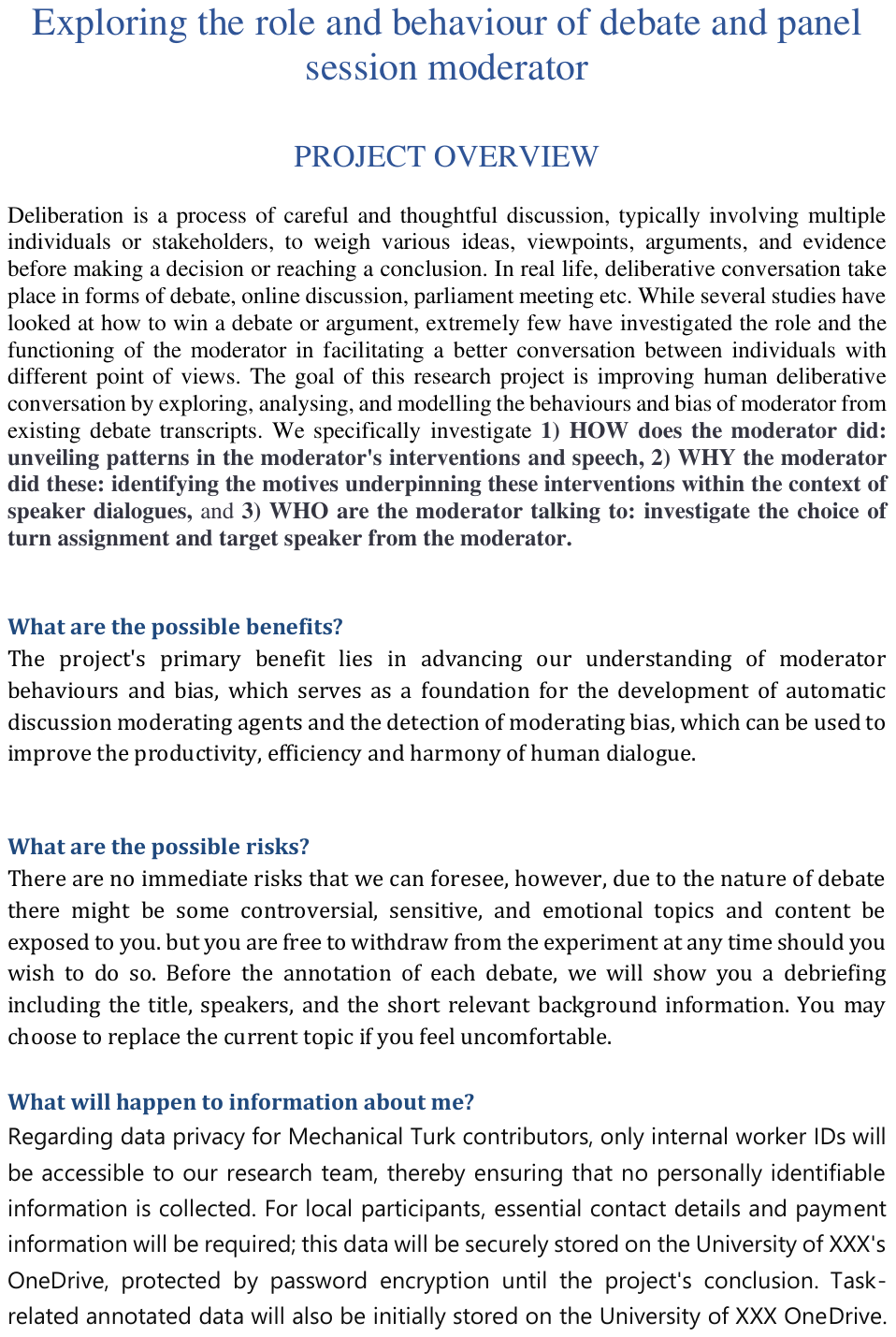}

\end{document}